\documentclass[10pt,twocolumn,letterpaper]{article}

\usepackage{cvpr}              % To produce the CAMERA-READY version
\usepackage[dvipsnames]{xcolor}

\definecolor{cvprblue}{rgb}{0.21,0.49,0.74}
\usepackage[pagebackref,breaklinks,colorlinks,citecolor=cvprblue]{hyperref}
\usepackage{multirow}
\usepackage{pifont}

\def\paperID{5924} % *** Enter the Paper ID here
\def\confName{CVPR}
\def\confYear{2024}

\title{Magic Tokens: Select Diverse Tokens for Multi-modal Object Re-Identification}

\author{{Pingping Zhang$^{1,2}$\thanks{Corresponding author}, Yuhao Wang$^{1}$, Yang Liu$^{1}$, Zhengzheng Tu$^{2,3}$ and Huchuan Lu$^{1}$}\\
{$^1$ School of Future Technology, School of Artificial Intelligence, Dalian University of Technology, China}\\
{$^2$ Anhui Provincial Key Laboratory of Multimodal Cognitive Computation, Anhui University, China}\\
{$^3$ School of Computer Science and Technology, Anhui University, China}\\
{\tt\small  \{zhpp,ly,lhchuan\}@dlut.edu.cn, 924973292@mail.dlut.edu.cn, zhengzhengahu@163.com}}

\begin{document}
\maketitle
\begin{abstract}
Single-modal object re-identification (ReID) faces great challenges in maintaining robustness within complex visual scenarios.
%Traditional 
In contrast, multi-modal object ReID utilizes complementary information from diverse modalities, showing great potentials for practical applications.
However, previous methods may be easily affected by irrelevant backgrounds and usually ignore the modality gaps.
To address above issues, we propose a novel learning framework named \textbf{EDITOR} to select diverse tokens from vision Transformers for multi-modal object ReID.
We begin with a shared vision Transformer to extract tokenized features from different input modalities.
Then, we introduce a Spatial-Frequency Token Selection (SFTS) module to adaptively select object-centric tokens with both spatial and frequency information.
Afterwards, we employ a Hierarchical Masked Aggregation (HMA) module to facilitate feature interactions within and across modalities.
Finally, to further reduce the effect of backgrounds, we propose a Background Consistency Constraint (BCC) and an Object-Centric Feature Refinement (OCFR).
They are formulated as two new loss functions, which improve the feature discrimination with background suppression.
As a result, our framework can generate more discriminative features for multi-modal object ReID.
Extensive experiments on three multi-modal ReID benchmarks verify the effectiveness of our methods.
The code is available at https://github.com/924973292/EDITOR.
\end{abstract} 
\section{Introduction}
\label{sec:intro}
\begin{figure}[t]
  \centering
  \includegraphics[width=1.0\linewidth]{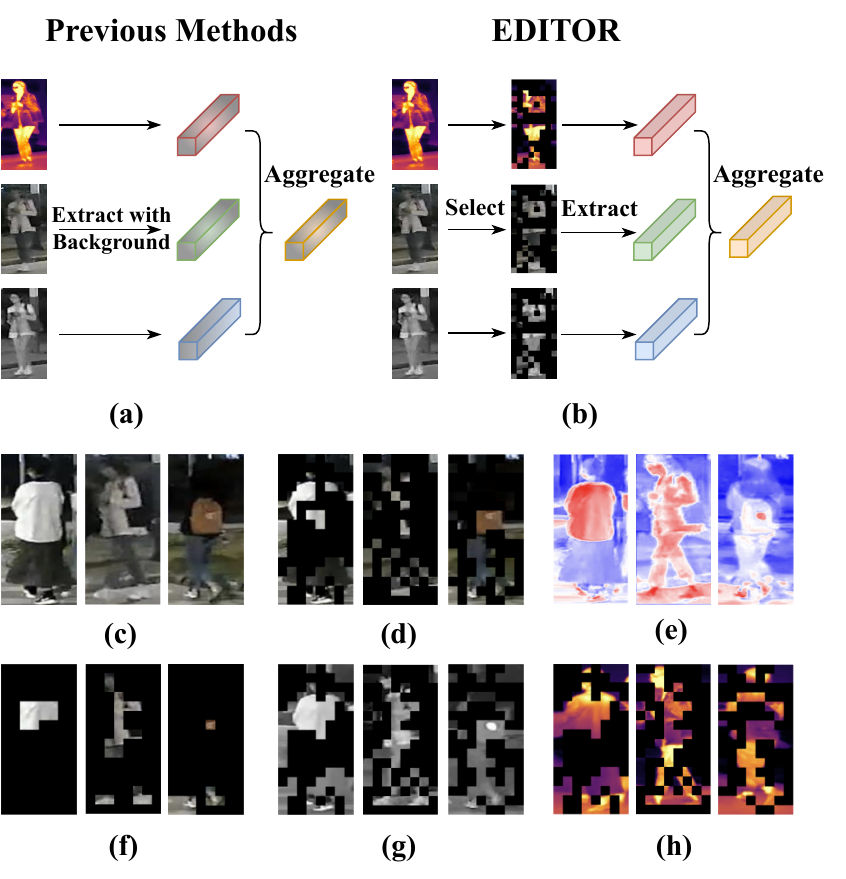}
  \vspace{-4mm}
  \caption{Comparison of different methods and token selections.
   (a) Framework of previous methods;
   (b) Framework of our proposed EDITOR;
   (c) RGB images;
   (d) Spatial-based token selection;
   (e) Multi-modal frequency transform;
   (f) Frequency-based token selection;
   (g) Selected tokens in the NIR modality;
   (h) Selected tokens in the TIR modality.}
   \label{fig:Introduction}
   \vspace{-4mm}
\end{figure}
Object re-identification (ReID) aims to retrieve specific objects (e.g., person, vehicle) across non-overlapping cameras.
Over the past few decades, object ReID has advanced significantly.
However, traditional object ReID with single-modal input encounters substantial challenges~\cite{li2020multi}, particularly in complex visual scenarios, such as extreme illumination, thick fog and low image resolution.
It can result in noticeable distortions in critical object regions, leading to disruptions during the retrieval process~\cite{zheng2021robust}.
Therefore, there has been a notable shift toward multi-modal approaches in recent years, capitalizing on diverse data sources to enhance the feature robustness for practical applications~\cite{zheng2021robust,wang2022interact,wang2023top}.
However, as illustrated in Fig.~\ref{fig:Introduction}, previous multi-modal ReID methods typically extract global features from all regions of images in different modalities and subsequently aggregate them.
Nevertheless, these methods present two key limitations:
(1) Within individual modalities, backgrounds introduce additional noise~\cite{tian2018eliminating}, especially in challenging visual scenarios.
(2) Across different modalities, backgrounds introduce overhead in reducing modality gaps, which may amplify the difficulty in aggregating features~\cite{huang2019sbsgan}.
Hence, our method prioritizes the selection of object-centric information, aiming to preserve the diverse features of different modalities while minimizing background interference.

To address above issues, we propose a novel feature learning framework named EDITOR to select diverse tokens for multi-modal object ReID.
Our EDITOR comprises two key modules: Spatial-Frequency Token Selection (SFTS) and Hierarchical Masked Aggregation (HMA).
Technically, we begin with a shared vision Transformer (ViT)~\cite{dosovitskiy2020image} to extract tokenized features from different input modalities.
Then, SFTS employs a dual approach to select object-centric tokens from both spatial and frequency perspectives.
In the spatial-based token selection, we combine all spatial indices selected by various heads in multi-head self-attention~\cite{dosovitskiy2020image} within each modality.
Afterwards, we further combine the indices from different modalities to enhance the token diversity across modalities.
However, as shown in Fig.~\ref{fig:Introduction} (d), the spatial-based token selection may not fully capture all object-centric tokens.
Therefore, we incorporate a frequency-based token selection to collaboratively extract the most salient tokens, as shown in Fig.~\ref{fig:Introduction} (e)-(f).
With the selected tokens, we introduce HMA to effectively aggregate object-centric tokens within and across modalities.
To further reduce the effect of backgrounds, we propose a Background Consistency Constraint (BCC) and an Object-Centric Feature Refinement (OCFR).
They are formulated as two new loss functions, which improve the feature discrimination with background suppressions.
With the proposed modules, our framework can extract more discriminative features for multi-modal object ReID.
Experiments on the three multi-modal object ReID benchmarks, i.e., RGBNT201, RGBNT100 and MSVR310 demonstrate the effectiveness of our proposed EDITOR.

In summary, our contributions are as follows:
\begin{itemize}
  \item
  We introduce EDITOR, a novel feature learning framework for multi-modal object ReID.
  To our best knowledge, it is the first attempt to enhance multi-modal object ReID through object-centric token selection.
  \item
  We propose a Spatial-Frequency Token Selection (SFTS) module and a Hierarchical Masked Aggregation (HMA) module.
  These modules effectively facilitate the selection and aggregation of multi-modal tokenized features.
  \item
  We propose two new loss functions with a Background Consistency Constraint (BCC) and an Object-Centric Feature Refinement (OCFR) to improve the feature discrimination with background suppressions.
  \item
  Extensive experiments are performed on three multi-modal object ReID benchmarks.
  The results fully validate the effectiveness of our proposed methods.
\end{itemize} 
\section{Related Work}
\label{sec:related}
\subsection{Single-modal Object ReID}
Single-modal object ReID extracts discriminative features from single-modality inputs, such as RGB, Near Infrared (NIR), Thermal Infrared (TIR), or depth images.
Most of existing object ReID methods are based on Convolutional Neural Networks (CNNs) or Transformers~\cite{vaswani2017attention}.
Regarding CNN-based methods, PCB \cite{sun2018beyond} and MGN \cite{wang2018learning} employ a part-based image partitioning approach to extract features at multiple levels of granularity.
In addition, with a unified aggregation gate mechanism, OSNet \cite{zhou2019omni} dynamically fuses features across omni-scales.
DMML \cite{chen2019deep} offers a meta-level view of metric learning, demonstrating the alignment of softmax and triplet losses in the meta space.
Circle loss \cite{sun2020circle} introduces a novel approach to re-weight similarity scores and achieve a more flexible optimization.
AGW \cite{ye2021deep} extracts fine-grained features with non-local attention mechanisms.
However, CNN-based methods~\cite{8513884,Liu_2021_CVPR,Wang_2021_ICCV,liu2023video} may not be sufficiently robust in complex scenarios due to their limited receptive field.
Drawing inspiration from the success of ViT~\cite{dosovitskiy2020image}, the first pure Transformer-based method named TransReID~\cite{he2021transreid} is proposed with the adaptive modeling of image patches, yielding competitive results.
Furthermore, AAformer~\cite{zhu2021aaformer} introduces an automated alignment strategy to extract local features.
DCAL~\cite{zhu2022dual} proposes a dual cross-attention method, which enhances self-attention with global-local and pairwise cross-attentions.
PHA~\cite{zhang2023pha} improves ViTs by enhancing high-frequency feature representations through a patch-wise contrastive loss.
Moreover, a multitude of Transformer-based approaches, as presented in works~\cite{zhu2022pass,liu2023deeply,yan2023learning,zhang2021hat,yu2023tfclip,wang2022nformer,lu2023learning,liu2021video}, showcase their benefits in object ReID.
Nevertheless, these approaches rely on single-modal input, which offer limited representation capabilities, especially in complex scenarios.
In contrast, our proposed EDITOR integrates diverse modalities and leverages token selections, enabling to capture more fine-grained features in a variety of scenarios.
\begin{figure*}[t]
  \centering
    \resizebox{1\textwidth}{!}
	{
  \includegraphics[width=1.\linewidth]{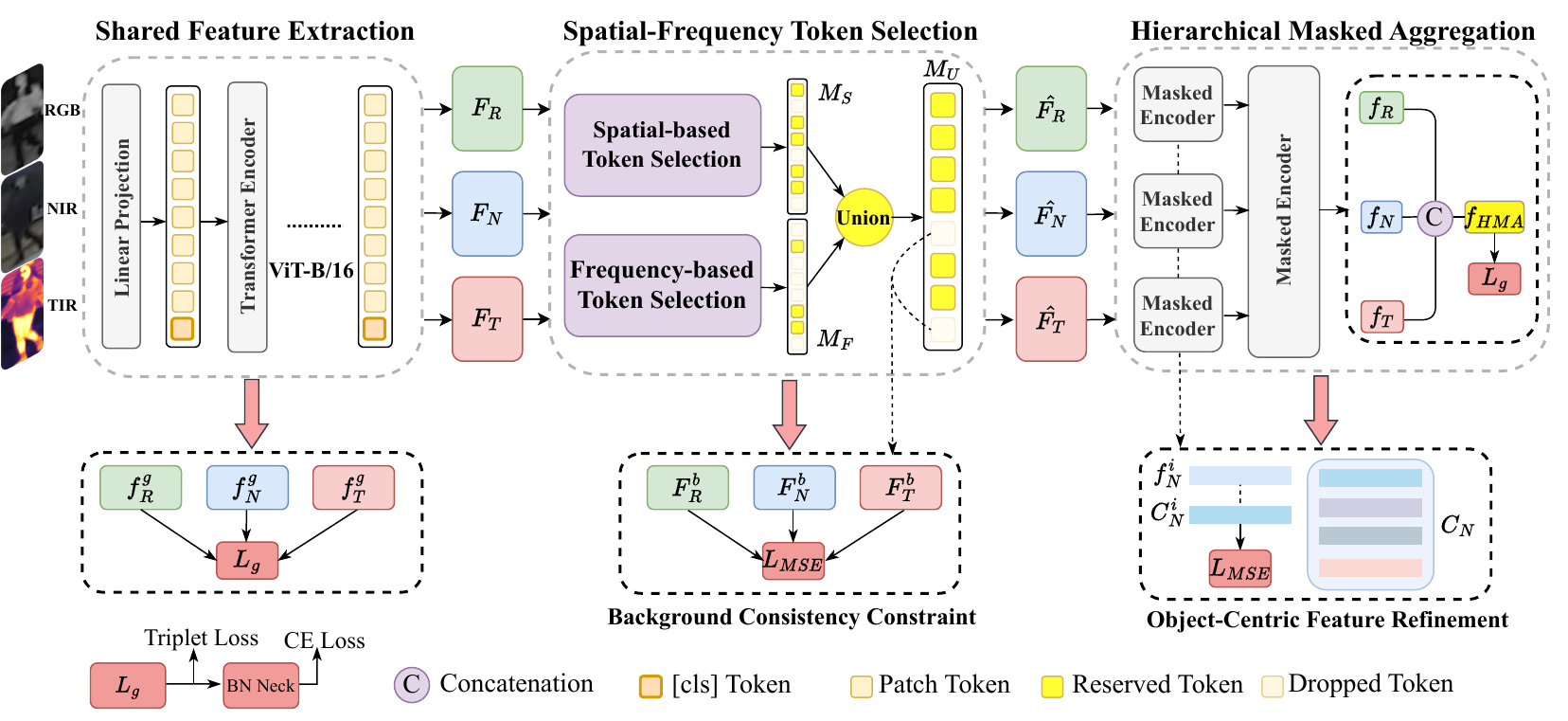}
  }
   \caption{An illustration of our proposed EDITOR.
   First, features from different input modalities are extracted by using the shared ViT-B/16 backbone.
   Then, a Spatial-Frequency Token Selection (SFTS) is utilized to select diverse tokens with object-centric features.
   Meanwhile, the Background Consistency Constraint (BCC) loss is designed for stabilizing the selection process.
   After that, a Hierarchical Masked Aggregation (HMA) is grafted to aggregate the selected tokens.
   Finally, combined with the Object-Centric Feature Refinement (OCFR) loss, the whole framework can obtain more discriminative features for multi-modal object ReID.
   }
  \label{fig:Overall}
  \vspace{-4mm}
\end{figure*}
\subsection{Multi-modal Object ReID}
Aggregating robust representations from multi-modal data has attracted considerable attention in recent years.
In the field of multi-modal person ReID, Zheng \emph{et al.}~\cite{zheng2021robust} first design a PFNet to learn robust features with progressive fusion.
Wang \emph{et al.}~\cite{wang2022interact} advance the field further with their IEEE framework, which employs three learning strategies to enhance modality-specific representations.
Then, Zheng \emph{et al.}~\cite{zheng2023dynamic} introduce the pixel-level reconstruction to address the modal-missing problem.
For multi-modal vehicle ReID, Li \emph{et al.}~\cite{li2020multi} propose a HAMNet to fuse different modal features with a heterogeneous score coherence loss.
Then, Zheng \emph{et al.}~\cite{zheng2022multi} reduce the discrepancies from sample and modality aspects.
From the perspective of generating modalities, Guo \emph{et al.}~\cite{guo2022generative} propose a GAFNet to fuse the multiple data sources.
He \emph{et al.}~\cite{he2023graph} propose a GPFNet to adaptively fuse multi-modal features with graph learning.
With Transformers, Pan \emph{et al.}~\cite{pan2023progressively} introduce a PHT, employing a feature hybrid mechanism to balance modal-specific and modal-shared information.
Jennifer \emph{et al.}~\cite{crawford2023unicat} provide a UniCat by analyzing the issue of modality laziness.
Very recently, Wang \emph{et al.}~\cite{wang2023top} propose a novel token permutation mechanism for robust multi-modal object ReID.
While contributing to the multi-modal object ReID, they commonly overlook the influence of irrelevant backgrounds on the aggregation of features across different modalities.
In contrast, our proposed EDITOR explicitly addresses the influence of irrelevant backgrounds on multi-modal feature aggregation.
Our approach effectively identifies critical regions within each modality while fostering inter-modal collaboration.
Furthermore, the incorporation of BCC and OCRF losses, along with the innovative SFTS and HMA modules, distinguishes our work as a promising avenue for improved performance in complex scenarios.
\subsection{Token Selection in Transformer}
With the increasing adoption of Transformers~\cite{liu2021swin,radford2021learning,kirillov2023segment}, token selection has gained significant attention~\cite{he2022transfg,rao2021dynamicvit,yang2023top,bolya2022token,fayyaz2022adaptive,liu2022ts2,haurum2023tokens,marin2021token}, due to its ability to focus on essential objects and reduce computational overhead.
In vision tasks, such as ReID, where fine-grained features are crucial, the extraction of key regions becomes particularly important.
For example, TransFG~\cite{he2022transfg} utilizes the multi-head self-attention of ViT to select representative local patches, achieving outstanding performance in fine-grained classification tasks.
DynamicViT~\cite{rao2021dynamicvit} employs gating mechanisms to dynamically accelerate both training and inference.
TVTR~\cite{yang2023top} extends token selection to cross-modal ReID, aligning features by selecting the top-K salient tokens.
However, our method differs from them in the following ways:
(1) Our selection is \textbf{instance-level}, where for different input images, the model dynamically selects different numbers of object-centirc tokens.
Unlike previous methods, which specify the fixed top-K local regions for feature aggregation, our approach allows the model to adapt more flexibly to various inputs.
(2) Previous methods do not consider the impact of \textbf{distracted backgrounds} during the early selection process.
With our proposed losses, we effectively stabilize the selection process, achieving dynamic distribution alignments.
Thus, we provide a more flexible framework, ultimately enhancing ReID performance in complex scenarios. 
\section{Proposed Method}
\label{sec:methods}
As illustrated in Fig.~\ref{fig:Overall}, our proposed EDITOR comprises three key components: Shared Feature Extraction, Spatial-Frequency Token Selection (SFTS) and Hierarchical Masked Aggregation (HMA).
In addition, we incorporate the Background Consistency Constraint (BCC) and Object-Centric Feature Refinement (OCFR) to further reduce the effect of irrelevant backgrounds.
We will describe these key modules in the following subsections.
\subsection{Shared Feature Extraction}
To extract multi-modal features while reducing the model parameters, we deploy a shared vision Transformer (ViT) for multi-modal inputs.
Without loss of generality, for the RGB, NIR and TIR modalities, the multi-modal tokenized features can be expressed as:
\begin{equation} F_{R} = \mathrm{ViT}\left(I_{R}\right),
  F_{N} = \mathrm{ViT}\left(I_{N}\right),
  F_{T} = \mathrm{ViT}\left(I_{T}\right),
\end{equation}
\noindent where $I_R$, $I_N$ and $I_T$ represent the input RGB, NIR and TIR images, respectively.
The tokenized features $F_R$, $F_N$ and $F_T$, each of which has a shape of $\mathbb{R}^{D \times (N_{p}+1)}$, are extracted from the last layer of $\mathrm{ViT}$.
Here, we follow previous works and employ additional learnable class tokens $f_R^{g}$, $f_N^{g}$ and $f_T^{g}$ for corresponding modalities.
$N_{p}$ means the number of patch tokens while $D$ is the embedding dimension.
\subsection{Spatial-Frequency Token Selection}
\begin{figure}[t]
  \centering
  \includegraphics[width=0.8\linewidth]{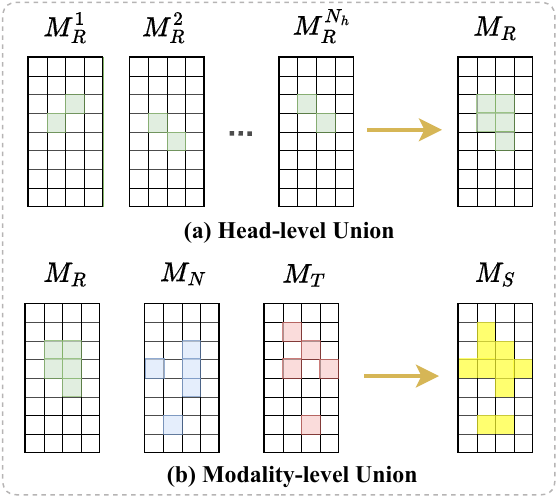}
  \vspace{-1mm}
  \caption{Illustration of spatial-based token selection.}
  \label{fig:spatial}
  \vspace{-4mm}
\end{figure}
To preserve diverse information within and across modalities while eliminating the influence of irrelevant backgrounds, we propose the Spatial-Frequency Token Selection (SFTS) module.
It consists of spatial-based token selection and frequency-based token selection.
Through the collaboration of these two selection methods, our EDITOR can focus on the critical regions of the object.
\\
\textbf{Spatial-based Token Selection.}
As shown in Fig. \ref{fig:spatial}, our spatial-based token selection is enhanced by both head-level union and modality-level union.
This kind of combinations facilitate a dynamic selection of instance-level tokens, preserving diverse information across different input modalities.
Technically, the spatial-based token selection takes tokens from the three modalities and ultimately produces a selection mask $M_{S}$ for all three modalities.
Taking the RGB modality as an example, assuming there are a total of $K$ layers in the backbone network, and there are $N_{h}$ heads in the self-attention layer, the attention weights of the $k$-th layer for the RGB modality can be represented as follows:
\begin{equation}
  a_{k}^{i}=\left[a_{k}^{i_{0}} , a_{k}^{i_{1}}, a_{k}^{i_{2}}, \cdots, a_{k}^{i_{N_{p}}}\right], \quad i \in 1,2, \cdots, N_{h},
\end{equation}
where $a_{k}^{i}$ is the attention weight in the $i$-th head of the $k$-th layer.
$a_{k}^{i_{0}}$ is the corresponding weight of the class token.
Thus, the attention weights of all layers are organized as:
\begin{equation}
  A_{k} = \left[a_{k}^{1}, a_{k}^{2}, a_{k}^{3}, \cdots, a_{k}^{N_{h}}\right], \quad k \in 1,2, \cdots, K.
\end{equation}
To further concentrate attention on objects, we follow~\cite{he2022transfg} to integrate attention weights from all the preceding layers.
Specifically, the attention score is iteratively computed through a matrix multiplication in the following manner:
\begin{equation}
A_{score}=\prod_{k=1}^{K} A_{k}.
\end{equation}
Here, $A_{score}$ represents the comprehensive relationships between patches.
Then, we extract the weights associated with the class token $a_{score}^{i_{0}}$ from each head in $A_{score}$.
For each head, we retain the crucial tokens and generate the mask $M_R^i$.
This process can be formalized as:
\begin{equation}
M_R^i = \mathrm{Mask}(\mathrm{Top_{s}}(a_{score}^{i_{0}})),\quad i \in 1,2, \cdots, N_{h},
\end{equation}
where $\mathrm{Mask}$ represents transforming the selected tokens into a mask form, and $\mathrm{Top_{s}}$  retains the top $s$ important tokens ($s\in N^{+}$).
In multi-head self-attention, different heads focus on different aspects.
To capture more details within modality, we employ \textbf{head-level union} to combine the selected tokens from different heads.
Finally, we obtain the mask of RGB modality $M_{R}$, which can be formulated as:
\begin{equation}
M_{R} = \bigcup_{i=1}^{N_h} M_R^i.
\end{equation}
For other modalities, we execute similar operations as:
\begin{equation} M_{N} = \bigcup_{i=1}^{N_h} M_N^i,
M_{T} = \bigcup_{i=1}^{N_h} M_T^i.
\end{equation}
Based on above operations, we complete the selection process within individual modalities.
As a result, each modality chooses tokens that focus on objects and eliminate most background interferences.
However, the significant variations across modalities lead to challenges when directly aggregating these tokenized features.
To address this issue and facilitate the modality complementary, we further introduce \textbf{modality-level union}.
It can be expressed as follows:
\begin{equation}
  M_{S} = M_{R} \cup M_{N} \cup M_{T},
  \end{equation}
where $M_{S}$ means the final mask from spatial-based token selection.
Through the head-level union and modality-level union, we achieve an instance-level token selection strategy, providing diverse tokenized features for modal fusion.
\\
\textbf{Frequency-based Token Selection.}
\begin{figure}[t]
  \centering
  \includegraphics[width=0.8\linewidth]{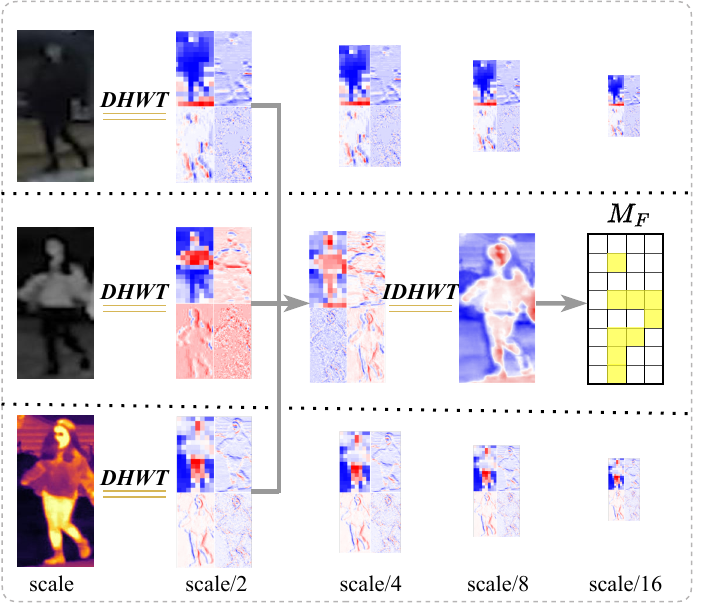}
  \vspace{-1mm}
  \caption{Illustration of frequency-based token selection.}
  \label{fig:frequency}
  \vspace{-4mm}
 \end{figure}
As shown in Fig.~\ref{fig:Introduction} (d), the spatial-based token selection may result in the neglect of some salient tokens.
Considering the frequency information could provide a structural perception of images, we introduce a frequency-based token selection to mine more salient tokens.
As shown in Fig.~\ref{fig:frequency}, we apply the Discrete Haar Wavelet Transform (DHWT)~\cite{mallat1989theory} to the images of different modalities.
Taking the RGB modality as an example, we obtain four frequency components:
\begin{equation}
  I_{R}^{ll},I_{R}^{lh},I_{R}^{hl},I_{R}^{hh} = \mathrm{DHWT}(I_{R}),
\end{equation}
where $I_{R}^{ll}$ is the low-frequency component, while the other three terms are the high-frequency components.
The same operations are carried out for other modalities.
It can be observed that the decomposition results of different modalities exhibit significant frequency differences.
Technically, we first sum up the decomposition results from different modalities at each scale.
Then, we perform the Inverse Discrete Haar Wavelet Transform (IDHWT) with the summed results to obtain the inverse transformed image.
As shown in the middle of Fig.~\ref{fig:frequency}, the inverse transformed image highlights salient regions.
Finally, we use a sliding window to count the pixel values within each patch.
The tokens of top $f$ values are selected as the frequency-based selection result, denoted as $M_{F}$.
As a result, by uniting spatial mask $M_{S}$ and frequency mask $M_{F}$, we obtain the final mask $M_{U}$.
\\
\textbf{Background Consistency Constraint.}
Most of previous methods~\cite{he2022transfg,yang2023top} directly discard non-selected tokens, treating them as backgrounds.
However, this may lead to the loss of important information.
Therefore, we step further to impose consistency constraints on these non-selected tokens from different modalities.
Formally, the background mask $M_{B} \in \mathbb{R}^{N_{p}}$ is defined as:
\begin{equation}
  M_{B} = 1 - M_{U}.
\end{equation}
Then the background tokens from different modalities can be represented as follows:
\begin{equation}
  F_{R}^b = F_{R}^{p} \odot M_{B},
  F_{N}^b = F_{N}^{p} \odot M_{B},
  F_{T}^b = F_{T}^{p} \odot M_{B},
\end{equation}
where $F_{R}^{p}$, $F_{N}^{p}$ and $F_{T}^{p}$ represents the patch features of $F_{R}$, $F_{N}$ and $F_{T}$, respectively.
$M_{B}$ is the background indices.
$\odot$ denotes the element-wise multiplication.
Then, the background tokens from paired modalities are constrained by the Mean Squared Error (MSE) loss:
\vspace{-3mm}
\begin{equation}
\mathcal{L}_{\mathrm{R2N}} = \sum_{i=1}^{N_{r}}{||F_{R}^b - F_{N}^b||_2^2},
\end{equation}
\vspace{-3mm}
\begin{equation}
\mathcal{L}_{\mathrm{R2T}} = \sum_{i=1}^{N_{r}}{||F_{R}^b - F_{T}^b||_2^2},
\end{equation}
\vspace{-3mm}
\begin{equation}
\mathcal{L}_{\mathrm{N2T}} = \sum_{i=1}^{N_{r}}{||F_{N}^b - F_{T}^b||_2^2}.
\end{equation}
The final consistency constraint loss can be formulated as:
\begin{equation}
\mathcal{L}_{BCC} = \frac{1}{N_{r}}(\mathcal{L}_{\mathrm{R2N}} + \mathcal{L}_{\mathrm{R2T}}+ \mathcal{L}_{\mathrm{N2T}}),
\end{equation}
where $N_r$ is the number of reserved tokens.
With $\mathcal{L}_{BCC}$, we can achieve dynamic alignments of backgrounds and stabilize the token selection process.
\subsection{Hierarchical Masked Aggregation}
For enhancing the feature robustness, we introduce the Hierarchical Masked Aggregation (HMA) to effectively aggregate selected diverse tokens from different modalities.
More specifically, the HMA consists of independent aggregation and collaborative aggregation.
In the independent aggregation stage, each modality interacts with its selected tokens, highlighting specific regions and improving the feature discrimination.
In the collaborative aggregation stage, tokens from all modalities interact with each other, facilitating the exchange and fusion of multi-modal information.
\\
\textbf{Independent Aggregation.}
Without loss of generality, taking the RGB modality as an example, we first concatenate $f_{R}^{g}$ with selected tokens to form $\hat{F_{R}} \in \mathbb{R}^{D\times(N_{p}+1)}$.
Then, it is fed into a masked encoder for feature interaction:
\begin{equation}
    {\overline{F}_{R}} = \varTheta(\hat{F_{R}}),
    \hat{F_{R}} = [f_{R}^{g},F_{R}^{p} \odot M_{U}],
\end{equation}
where $[\cdot]$ is the concatenation operation.
${\overline{F}_{R}} \in \mathbb{R}^{D\times(N_{p}+1)}$ represents the aggregated features.
The masked encoder $\varTheta$ is essentially a Transformer block with a Multi-Head Self-Attention (MHSA)~\cite{dosovitskiy2020image} and a Feed-forward Neural Network (FFN)~\cite{dosovitskiy2020image}.
Thus, we obtain tokenized features aligned with object-centric regions.
Other modalities are processed as:
\begin{equation}
    {\overline{F}_{T}} = \varTheta(\hat{F_{T}}),
    {\overline{F}_{N}} = \varTheta(\hat{F_{N}}).
    \end{equation}
As a result, global features of each modality will further focus on key tokens, obtaining object-centric features.
\\
\textbf{Collaborative Aggregation.}
To collaboratively aggregate tokens from different modalities, we first concatenate $\overline{F}_{R}$, $\overline{F}_{N}$ and $\overline{F}_{T}$ along the token dimension to form $\overline{F} \in \mathbb{R}^{D\times3(N_{p}+1)}$.
Then, it is fed into $\varTheta$ for feature interaction:
\begin{equation}
    {F}_{Agg} = \varTheta(\overline{F}),
    {\overline{F}} = [\overline{F}_{R},\overline{F}_{N},\overline{F}_{T}],
\end{equation}
where ${F}_{Agg} \in \mathbb{R}^{D\times3(N_{p}+1)}$ is the aggregated feature from different modalities.
Then, we extract class tokens $f_{R}$, $f_{N}$ and $f_{T}$ from ${F}_{Agg}$, and concatenate them as the output $f_{HMA} \in \mathbb{R}^{3D}$ of HMA.
With HMA, we achieve an adaptive token aggregation within and across modalities, enhancing the discriminative ability of multi-modal features.
\\
\textbf{Object-Centric Feature Refinement.}
After suppressing background interferences, we propose the Object-Centric Feature Refinement to enhance the aggregation of intra-modal features.
As shown in the bottom right corner of Fig. \ref{fig:Overall}, we construct and update the feature center for the $i$-th ID.
Without loss of generality, taking the NIR modality as an example, this is achieved by first computing the averaged feature belonging to the $i$-th ID in the current mini-batch:
\begin{equation}
  {f_{N}^{i}} = \frac{1}{P} \sum_{y_i=i} (\overline{F}_{N}^{cls}[\mathrm{label} = y_i]),
  \end{equation}
where $y_i$ is the label of the current feature, and $P$ is the number of instances in each ID in a mini-batch.
$\overline{F}_{N}^{cls}$ is the class token of $\overline{F}_{N}$.
Then, the updated center is:
\begin{equation}
  C_{N}^{i}|_{iter} := \alpha f_{N}^{i} + (1-\alpha) C_{N}^{i}|_{iter-1},
  \end{equation}
where $\alpha$  is the exponential decay rate, and $iter$ denotes the current iteration.
Furthermore, by using a MSE loss, we ensure that features belonging to the same ID are pulled closer to the ID center.
This can be represented as follows:
\begin{equation}
  \mathcal{L}_{N} = \frac{1}{B} \sum_{i}\sum_{y_i = i}{||\overline{F}_{N}^{cls}[\mathrm{label} = y_i] - C_{N}^{i}||_2^2},
  \end{equation}
where $B$ represents the batch size.
Similarly, the features from the RGB and TIR modalities will align with their respective centers, resulting in the following loss:
\begin{equation}
\mathcal{L}_{OCFR} = \mathcal{L}_{R} + \mathcal{L}_{N} + \mathcal{L}_{T}.
\end{equation}
\subsection{Objective Function}
As illustrated in Fig.~\ref{fig:Overall}, our objective function comprises four components: losses for the ViT backbone, HMA, BCC and OCFR.
For the backbone and HMA, they are both supervised by the label smoothing cross-entropy loss~\cite{szegedy2016rethinking} and triplet loss~\cite{hermans2017defense} with equal weights:
\begin{equation}
  \mathcal{L}_{g} = \mathcal{L}_{ce} + \mathcal{L}_{tri}.
\end{equation}
Finally, the total loss for our framework can be defined as:
\begin{equation}
\begin{split}
\mathcal{L}_{total} = \mathcal{L}_{g}^{ViT} + \mathcal{L}_{g}^{HMA} + \mathcal{L}_{BCC} + \mathcal{L}_{OCFR}.
\end{split}
\end{equation}
\vspace{-4mm} 
\section{Experiments}
\label{sec:experiments}
\begin{table*}[t]
    \caption{Performance comparison on three multi-modal object ReID benchmarks.
    The best and second results are in bold and underlined, respectively.
    *denotes Transformer-based methods, while the rest are CNN-based methods.
    Both single-modal and multi-modal methods are included.
    For the comparison between TOP-ReID and EDITOR, A and B means the AL setting and BL setting~\cite{wang2023top}, respectively.
    }
    %\vspace{-2mm}
    \begin{subtable}{.5\linewidth}
      \centering
      \caption{Comparison on RGBNT201.}
    \resizebox{!}{4.5cm}{
    \renewcommand\arraystretch{1.35}
    \setlength\tabcolsep{6pt}
    \begin{tabular}{cccccc}
            \hline
            \multicolumn{2}{c}{\multirow{2}{*}{Methods}}     &\multicolumn{4}{c}{RGBNT201}\\ \cline{3-6}
            & & mAP & R-1 & R-5 & R-10 \\
            \hline
            \multirow{6}{*}{Single}
            &MUDeep~\cite{qian2017multi} & 23.8 & 19.7 & 33.1 & 44.3 \\
            &HACNN~\cite{li2018harmonious} & 21.3 & 19.0 & 34.1 & 42.8 \\
            &MLFN~\cite{chang2018multi} & 26.1 & 24.2 & 35.9 & 44.1 \\
            &PCB~\cite{sun2018beyond}  & 32.8 & 28.1 & 37.4 & 46.9 \\
            &OSNet~\cite{zhou2019omni} & 25.4 & 22.3 & 35.1 & 44.7 \\
            &CAL~\cite{rao2021counterfactual}  & 27.6 & 24.3 & 36.5 & 45.7 \\
            \hline
            \multirow{6}{*}{Multi}
            & HAMNet~\cite{li2020multi}   & 27.7         & 26.3            & 41.5            & 51.7             \\
            & PFNet~\cite{zheng2021robust}    & 38.5         & 38.9            & 52.0              & 58.4             \\
            & IEEE~\cite{wang2022interact}     & 49.5         & 48.4            & 59.1           & 65.6             \\
            & DENet~\cite{zheng2023dynamic}    & 42.4         & 42.2            & 55.3            & 64.5            \\
            & UniCat{*}~\cite{crawford2023unicat}    & 57.0         & 55.7            & -            & -            \\
            & TOP-ReID (A){*}~\cite{wang2023top}  &\textbf{72.3} &\textbf{76.6} &\textbf{84.7} &\textbf{89.4}\\
            & TOP-ReID (B){*}~\cite{wang2023top}  &64.6 &64.6 &77.4 &82.4\\
            & \textbf{EDITOR (A)}{*} & \underline{66.5}       & 68.3           & 81.1        & 88.2             \\
            & \textbf{EDITOR (B)}{*} & 65.7       & \underline{68.8}           & \underline{82.5}         & \underline{89.1}             \\
            \hline
            \end{tabular}}
    \end{subtable}%
    \begin{subtable}{.5\linewidth}
      \centering
        \caption{Comparison on RGBNT100 and MSVR310.}
        \resizebox{!}{4.5cm}{
            \renewcommand\arraystretch{1.1}
            \setlength\tabcolsep{6pt}
            \begin{tabular}{cccccc}
                    \hline
                    \multicolumn{2}{c}{\multirow{2}{*}{Methods}} &  \multicolumn{2}{c}{RGBNT100} & \multicolumn{2}{c}{MSVR310} \\
                    \cline{3-6}
                    & & mAP & R-1 & mAP & R-1 \\
                    \hline
                    \multirow{9}{*}{Single}
                    &PCB~\cite{sun2018beyond}& 57.2 & 83.5 & 23.2 & 42.9 \\
                    &MGN~\cite{wang2018learning} & 58.1 & 83.1 & 26.2 & 44.3 \\
                    &DMML~\cite{chen2019deep}& 58.5 & 82.0 & 19.1 & 31.1 \\
                    &BoT~\cite{luo2019bag} & 78.0 & 95.1 & 23.5 & 38.4 \\
                    &OSNet~\cite{zhou2019omni}& 75.0 & 95.6 & 28.7 & 44.8 \\
                    &Circle Loss~\cite{sun2020circle}& 59.4 & 81.7 & 22.7 & 34.2 \\
                    &HRCN~\cite{zhao2021heterogeneous} & 67.1 & 91.8 & 23.4 & 44.2 \\
                    &AGW~\cite{ye2021deep} & 73.1 & 92.7 & 28.9 & 46.9 \\
                    &TransReID{*}~\cite{he2021transreid}& 75.6 & 92.9 & 18.4 & 29.6 \\
                    \hline
                    \multirow{7}{*}{Multi}
                    &HAMNet~\cite{li2020multi}  & 74.5 & 93.3 & 27.1 & 42.3 \\
                    &PFNet~\cite{zheng2021robust}& 68.1 & 94.1 & 23.5 & 37.4 \\
                    &GAFNet~\cite{guo2022generative} & 74.4 & 93.4 & - & - \\
                    &CCNet~\cite{zheng2022multi} & 77.2 & 96.3 & \underline{36.4} & \textbf{55.2} \\
                    &GraFT{*}~\cite{yin2023graft} &76.6 &94.3 &- &-\\
                    &GPFNet~\cite{he2023graph} & 75.0 & 94.5 & - & - \\
                    &PHT{*}~\cite{pan2023progressively}& 79.9 & 92.7 & - & - \\
                    &UniCat{*}~\cite{crawford2023unicat}    & 79.4         & 96.2  & -            & -            \\
                    &TOP-ReID (A){*}~\cite{wang2023top} &73.7 & 92.2 & 30.2 & 33.7 \\
                    &TOP-ReID (B){*}~\cite{wang2023top} &\underline{81.2} & \underline{96.4} & 35.9 & 44.6 \\
                    & \textbf{EDITOR (A){*}} & 79.8 & 93.9 & 35.8 & 43.1\\
                    & \textbf{EDITOR (B){*}} & \textbf{82.1} & \textbf{96.4} & \textbf{39.0} & \underline{49.3}\\
                    \hline
                    \end{tabular}
                    }
    \end{subtable}
    \label{Tab:Overall}
    \vspace{-4mm}
  \end{table*}
\subsection{Dataset and Evaluation Protocols}
To evaluate the performance of our method, we employ three multi-modal object ReID benchmarks.
More specifically, RGBNT201~\cite{zheng2021robust} is the first multi-modal person ReID dataset encompassing RGB, NIR, and TIR modalities.
RGBNT100~\cite{li2020multi} is a large-scale multi-modal vehicle ReID dataset.
MSVR310~\cite{zheng2022multi} is a small-scale multi-modal vehicle ReID dataset with complex visual scenarios.
As for evaluation metrics, we follow previous works and utilize the mean Average Precision (mAP) and Cumulative Matching Characteristics (CMC) at Rank-K ($K=1,5,10$).
\subsection{Implementation Details}
Our model is implemented by using the PyTorch toolbox.
Experiments are conducted on two NVIDIA A100 GPUs.
We employ pre-trained Transformers from the ImageNet classification dataset~\cite{deng2009imagenet} as our backbones.
For data processing, images are resized to 256$\times$128 for RGBNT201 and 128$\times$256 for RGBNT100/MSVR310.
During the training process, we employ random horizontal flipping, cropping, and erasing~\cite{zhong2020random} for data augmentation.
The mini-batch size is set to 128, containing 8 randomly selected object identities, and 16 images sampled for each identity.
To optimize our model, we use the Stochastic Gradient Descent (SGD) optimizer with a momentum of 0.9 and a weight decay of 0.0001.
The learning rate is initialized at 0.001 and follows a warmup strategy with a cosine decay.
In spatial-based token selection, $s$ is set to 2, while in frequency-based token selection, $f$ is set to 10.
For the OCFR, we set $\alpha$ to 0.8.
\subsection{Comparison with State-of-the-Art Methods}
We perform comparisons with state-of-the-art methods on three multi-modal ReID datasets.
Our method demonstrates competitive results compared with previous methods.
\\
\textbf{Multi-modal Person ReID.}
As presented in Tab.~\ref{Tab:Overall}, we compare EDITOR with both single-modal and multi-modal methods on RGBNT201.
In general, single-modal methods tend to exhibit lower performance.
Among the single-modal methods, PCB~\cite{sun2018beyond} stands out with an impressive mAP of 32.8\%, showcasing the effectiveness of its part-based matching strategy.
For multi-modal methods, TOP-ReID (B)~\cite{wang2023top} achieves a remarkable mAP of 64.6\%.
However, our EDITOR (B) with a mAP of 65.7\%, outperforms TOP-ReID (B), delivering an 1.1\% improvement.
Moreover, there is a noticeable improvement in the rank metrics, indicating the effectiveness of our method in addressing the challenges of multi-modal person ReID.
Although showing inferior performance than TOP-ReID (A), EDITOR (A) is more robust across different settings, potentially addressing the modality laziness problem~\cite{crawford2023unicat}.
Besides, EDITOR has fewer parameters than TOP-ReID, making it more efficient.
\\
\textbf{Multi-modal Vehicle ReID.}
As shown in Tab. \ref{Tab:Overall}, single-modal methods generally exhibit lower performance compared with multi-modal methods.
In single-modal methods, CNN-based methods like AGW~\cite{ye2021deep}, OSNet~\cite{zhou2019omni} and BoT~\cite{luo2019bag} consistently achieve better results across datasets.
While Transformer-based methods, such as TransReID~\cite{he2021transreid}, exhibit slightly inferior performance, especially on smaller datasets like MSVR310, where they lag behind CNN-based methods.
However, Transformer-based methods prove their effectiveness in integrating multi-modal data.
Specifically, TOP-ReID (B)~\cite{wang2023top} achieves a mAP of 81.2\% on RGBNT100.
Our EDITOR (B) surpasses TOP-ReID (B) on RGBNT100, demonstrating a 0.9\% higher mAP.
Notably, our improvement over TransReID on the smaller dataset MSVR310 highlights our model's resilience to over-fitting.
Meanwhile, our EDITOR (B) achieves a 2.6\% higher mAP than CCNet.
These results verify the effectiveness of our method in multi-modal vehicle ReID.
\subsection{Ablation Studies}
We conduct ablation studies on the RGBNT201 dataset to validate the proposed components.
Our baseline utilizes a ViT-B/16 with camera embeddings, supervised by $\mathcal{L}_{g}^{ViT}$.
%--------------------------------------------------------------------------------------------------------
\begin{table}[t]
    \centering
    \renewcommand\arraystretch{1.12}
    \setlength\tabcolsep{3pt}
    \caption{Performance comparison with different components.}
    \resizebox{0.42\textwidth}{!}
	{
    \begin{tabular}{c|cc|cc|cccc}
        \hline
        &\multicolumn{2}{c|}{Module} &\multicolumn{2}{c|}{Loss}       &\multicolumn{4}{c}{RGBNT201} \\
        & SFTS& HMA& BCC& OCFR      & mAP & R-1 & R-5 & R-10 \\
        \hline
        A & \ding{53}& \ding{53}& \ding{53}& \ding{53}& 54.0	&53.5	&70.2	&78.8\\
        B & \ding{53}& \ding{51}& \ding{53}& \ding{53}& 60.7 & 62.4 &77.2 & 83.6 \\
        C & \ding{51}& \ding{51}& \ding{53}& \ding{53}& 62.2 & 65.0 & 79.3 & 85.4 \\
        D & \ding{51}& \ding{51}& \ding{51}& \ding{53}& 65.2 &	65.9 & 82.2	&87.1\\
        E & \ding{51}& \ding{51}& \ding{53}& \ding{51}& 64.8 & 66.9 & 82.3 & 87.3 \\
        F & \ding{51}& \ding{51}& \ding{51}& \ding{51}& \textbf{65.7}        & \textbf{68.8}           & \textbf{82.5}         & \textbf{89.1}             \\
        \hline
    \end{tabular}
    }
    \label{tab:ablation}
\end{table}
\\
\textbf{Effect of Key Components.}
Tab. \ref{tab:ablation} shows the performance comparison with different components.
The model A is the baseline.
Model B incorporates HMA, resulting in a 6.7\% increase in mAP, demonstrating the effectiveness in aggregating multi-modal features.
Furthermore, Model C introduces SFTS, achieving further performance improvement through object-centric token selection.
The introduction of BCC effectively achieves dynamic alignments of multi-modal distributions, resulting in a 3\% mAP improvement compared with Model C.
Besides, Model E makes the feature distribution more compact, leading to robust improvements.
By integrating all components, our model achieves the optimal performance.
These results validate the effectiveness of our EDITOR in complex scenarios.
\\
\textbf{Effect of Modality Union vs. Separation.}
As shown in Tab.~\ref{tab:sepvsunion}, we verify the effect of using modality union.
The results show that modality union significantly improves the performance.
Selected tokens from different modalities vary significantly, potentially causing instability in the subsequent aggregation.
This is evident in the second row of Tab. \ref{tab:sepvsunion}.
Therefore, by establishing shared indices, our modality union enables a collaborative interaction among different modalities, providing a more stable aggregation.
\begin{table}[t]
    \centering
    \renewcommand\arraystretch{1.12}
    \setlength\tabcolsep{3.8pt}
    \caption{Comparison between modality union and separation. The BCC and OCFR losses are not added here.}
    \resizebox{0.35\textwidth}{!}
	{
    \begin{tabular}{cccccc}
    \hline
    \multicolumn{1}{c}{\multirow{2}{*}{Methods}}     &\multicolumn{4}{c}{RGBNT201}\\ \cline{2-5}
     & mAP & R-1 & R-5 & R-10 \\
    \hline
    w/o selection& 60.7         & 62.4            & 77.2            & 83.6             \\
    w/ separation   & 57.7	& 58.5	& 75.4	& 82.5   \\
     \textbf{w/ union} & \textbf{62.2}        & \textbf{65.0}           & \textbf{79.3}         & \textbf{85.4}             \\
    \hline
    \end{tabular}
    }
    \label{tab:sepvsunion}
\end{table}
\\
\textbf{Effect of Different Selection Methods.}
\begin{table}[t]
    \centering
    \renewcommand\arraystretch{1.12}
    \setlength\tabcolsep{4pt}
    \caption{Effect of different selection methods in SFTS.}
    \resizebox{0.45\textwidth}{!}
	{
    \begin{tabular}{cccc}
    \hline
    \multicolumn{1}{c}{\multirow{2}{*}{Selection Methods}} &\multicolumn{1}{c}{\multirow{1}{*}{Reserved Tokens}}    &\multicolumn{2}{c}{RGBNT201}\\ \cline{2-4}
    &Average number & mAP & R-1  \\
    \hline
     Modality  & 30.2 & 64.2         & 65.7               \\
     Spatial & 55.0 & 65.0         & 66.8            \\
     Frequency &55.0 & 64.1         & 65.3                   \\
     \textbf{Spatial+Frequency} &58.0 & \textbf{65.7}        & \textbf{68.8}                    \\
    \hline
    \end{tabular}
    }
    \label{tab:different_union}
\end{table}
In Tab.~\ref{tab:different_union}, we validate different selection methods in SFTS.
The first row is modality union, and the second row introduces head union, forming the complete spatial-based token selection.
The introduction of head union increases retained tokens, leading to better performance.
In contrast, frequency-based token selection shows inferior results.
The best results are achieved by combining them.
\subsection{Visualization}
\begin{figure}[t]
    \centering
    \resizebox{0.46\textwidth}{!}
	{
    \includegraphics[width=1.\linewidth]{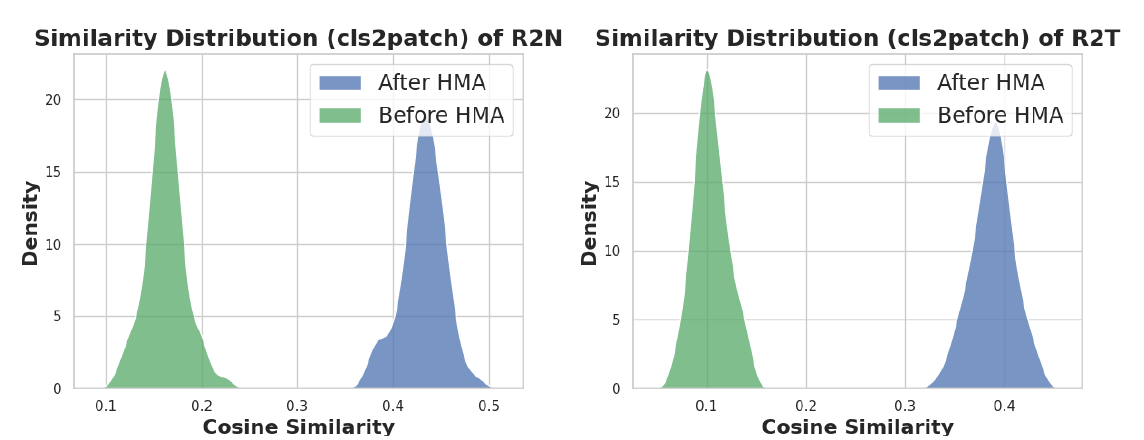}
    }
    \caption{Alignment visualization in HMA with RGB modality.}
    \label{fig:dis_HMF}
  \end{figure}
\begin{figure}[t]
    \centering
    \resizebox{0.47\textwidth}{!}
	{
    \includegraphics[width=1.0\linewidth,height=8.0cm]{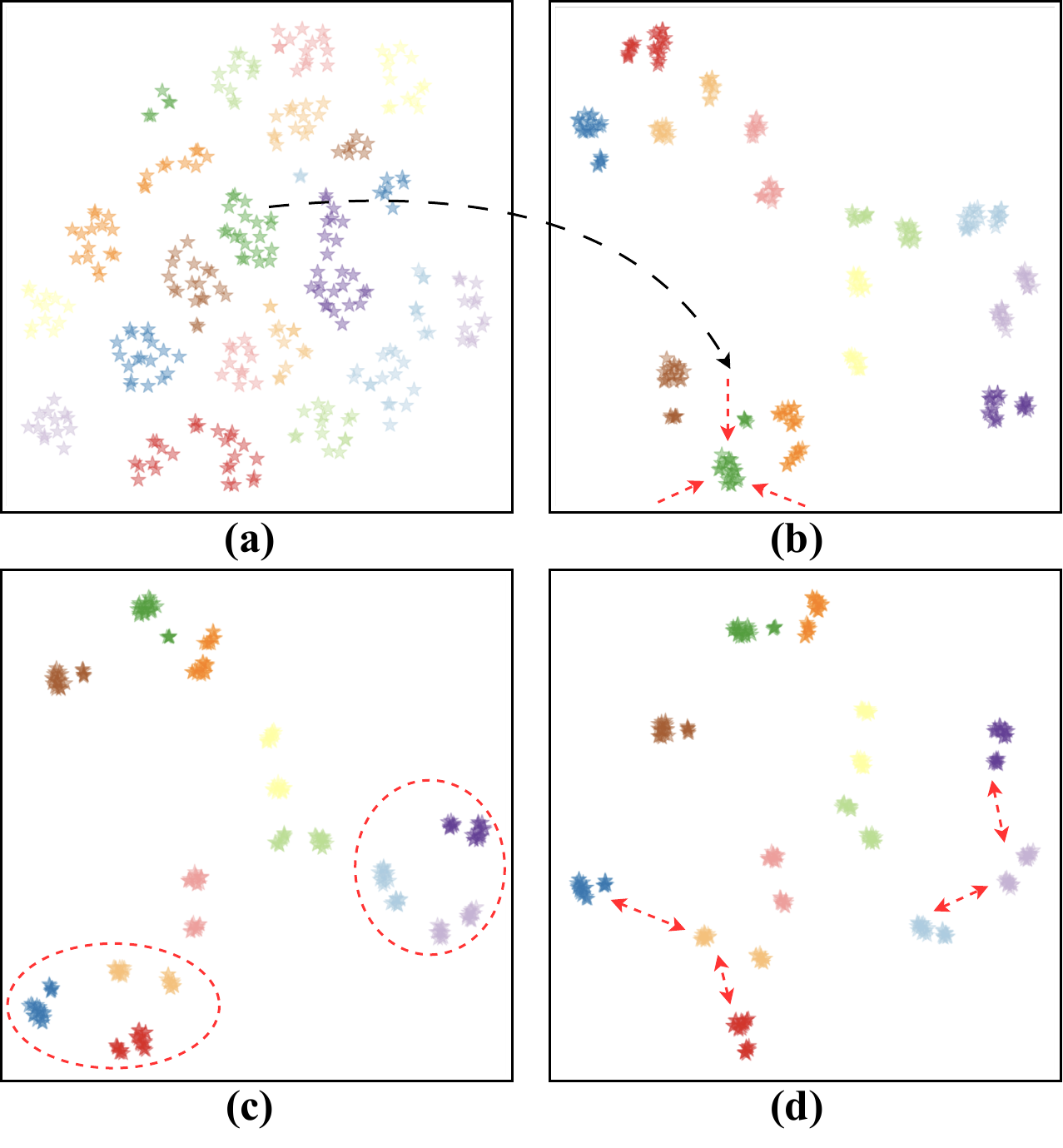}
    }
    \caption{Comparison of feature distributions with t-SNE~\cite{van2008visualizing}.
    Different colors represent different identities.
    (a) Baseline;
    (b) Baseline + SFTS + HMA;
    (c) Baseline + SFTS + HMA + OCFR;
    (d) Baseline + SFTS + HMA + OCFR + BCC.}
    \label{fig:dis}
    \vspace{-2mm}
  \end{figure}
\textbf{Feature Alignment of HMA.}
In Fig. \ref{fig:dis_HMF}, we measure the cosine similarity between class tokens of different modalities before and after HMA.
The results show that, after HMA, the class token of the RGB modality effectively aligns with patch tokens of other modalities.
Similar results can be observed for other modalities, confirming the effectiveness of HMA on feature alignment and aggregation.
More visualizations are provided in the supplementary material.
\\
\textbf{Feature Distributions.}
Fig. \ref{fig:dis} shows the feature distributions with different components.
When comparing Fig. \ref{fig:dis}(a) and Fig. \ref{fig:dis}(b), one can observe that SFTS and HMA can pull features to their ID clusters and increase gaps between different IDs.
As shown in Fig. \ref{fig:dis}(c), with OCFR, it obtains more compact features within the same ID, effectively enhancing feature distinctiveness.
Finally, with BCC, it enlarges gaps between different IDs.
These visualizations vividly verify the effectiveness of different modules. 
\section{Conclusion}
\label{sec:conclusion}
In this work, we propose EDITOR, a novel feature learning framework that selects diverse tokens from vision Transformers for multi-modal object ReID.
Our framework integrates Spatial-Frequency Token Selection (SFTS) and Hierarchical Masked Aggregation (HMA), which select and aggregate multi-modal features, respectively.
%effectively
To reduce the effect of backgrounds, we introduce Background Consistency Constraint (BCC) and Object-Centric Feature Refinement (OCFR) losses.
Extensive experiments on three benchmarks validate the effectiveness of our method.
\\
\small
\textbf{Acknowledgements.}
This work was supported in part by the National Natural Science Foundation of China (No.62101092), Open Project of Anhui Provincial Key Laboratory of Multimodal Cognitive Computation, Anhui University (No.MMC202102) and Fundamental Research Funds for the Central Universities (No.DUT22QN228 and No.DUT23BK050).
\vspace*{-4mm} 
{
    \small
    \bibliographystyle{ieeenat_fullname}
    \bibliography{main}
}

\setcounter{page}{1}
\maketitlesupplementary
\section{Introduction}
The supplementary material validates the effectiveness of our EDITOR with additional evidences.
We extend our ablation experiments to vehicle datasets, providing more visualization results.
More specifically, the experiment section provides more insights and explore the impact of various hyper-parameters.
The visualization section shows the selection effects with different kinds of objects, such as person and vehicle.
In conclusion, the supplementary material provides a comprehensive exploration of EDITOR's effectiveness, extending its applicability beyond person-centric scenarios to vehicles.
\section{Experiments}
\subsection{More Ablations on Multi-modal Person ReID}
\begin{figure}[t]
    \centering
    \resizebox{0.5\textwidth}{!}
	{
    \includegraphics[width=1.\linewidth]{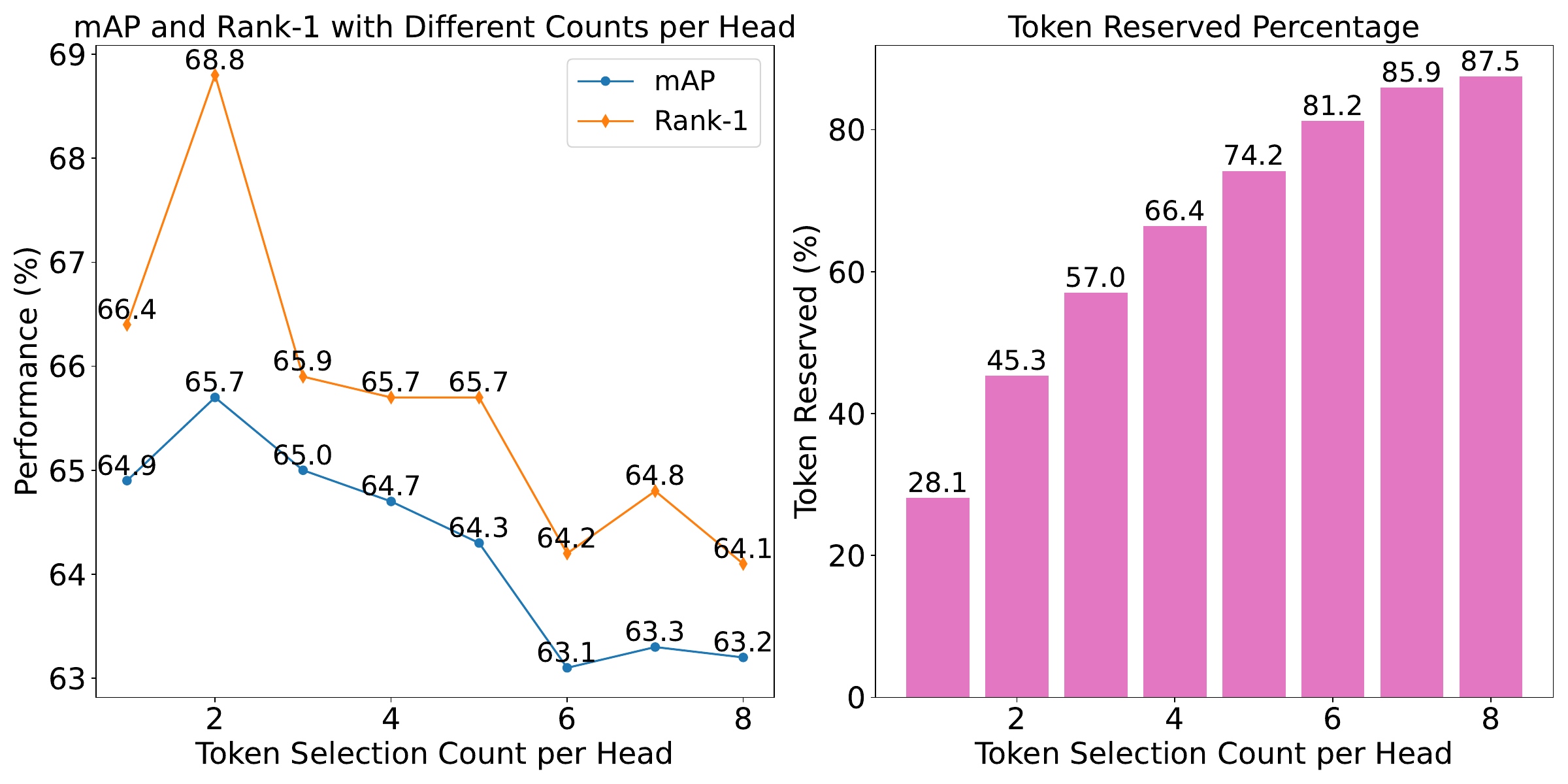}
    }
    \vspace{-2mm}
     \caption{Performance comparison of spatial-based selection.}
     \vspace{-4mm}
     \label{fig:head}
\end{figure}

\textbf{Effect of Patch Features in HMA.}
In Tab. \ref{tab:patch}, the first row indicates the absence of using averaged patches.
In this scenario, after HMA, each modality's class token is concatenated to form the final retrieval representation.
Conversely, the second row signifies the usage of averaged patches, where each modality's global feature is concatenated with the average token of the selected local features, followed by global supervision.
The comparison highlights the importance of local features, providing fine-grained details.
\begin{table}[h]
  \centering
  \renewcommand\arraystretch{1.12}
  \setlength\tabcolsep{3.8pt}
  \begin{tabular}{cccccc}
    \hline
  \multicolumn{1}{c}{\multirow{2}{*}{Methods}}     &\multicolumn{4}{c}{RGBNT201}\\ \cline{2-5}
   & mAP & R-1 & R-5 & R-10 \\
   \hline
  w/o averaged patches  & 61.0         & 60.8            & 75.2            & 82.4             \\
  w/ averaged patches   & 65.7	& 68.8	& 82.5	& 89.1   \\
  \hline
  \end{tabular}
  \caption{Effect of local features.}
  \vspace{-2mm}
  \label{tab:patch}
\end{table}
\begin{figure}[t]
  \centering
  \includegraphics[width=1.000\linewidth]{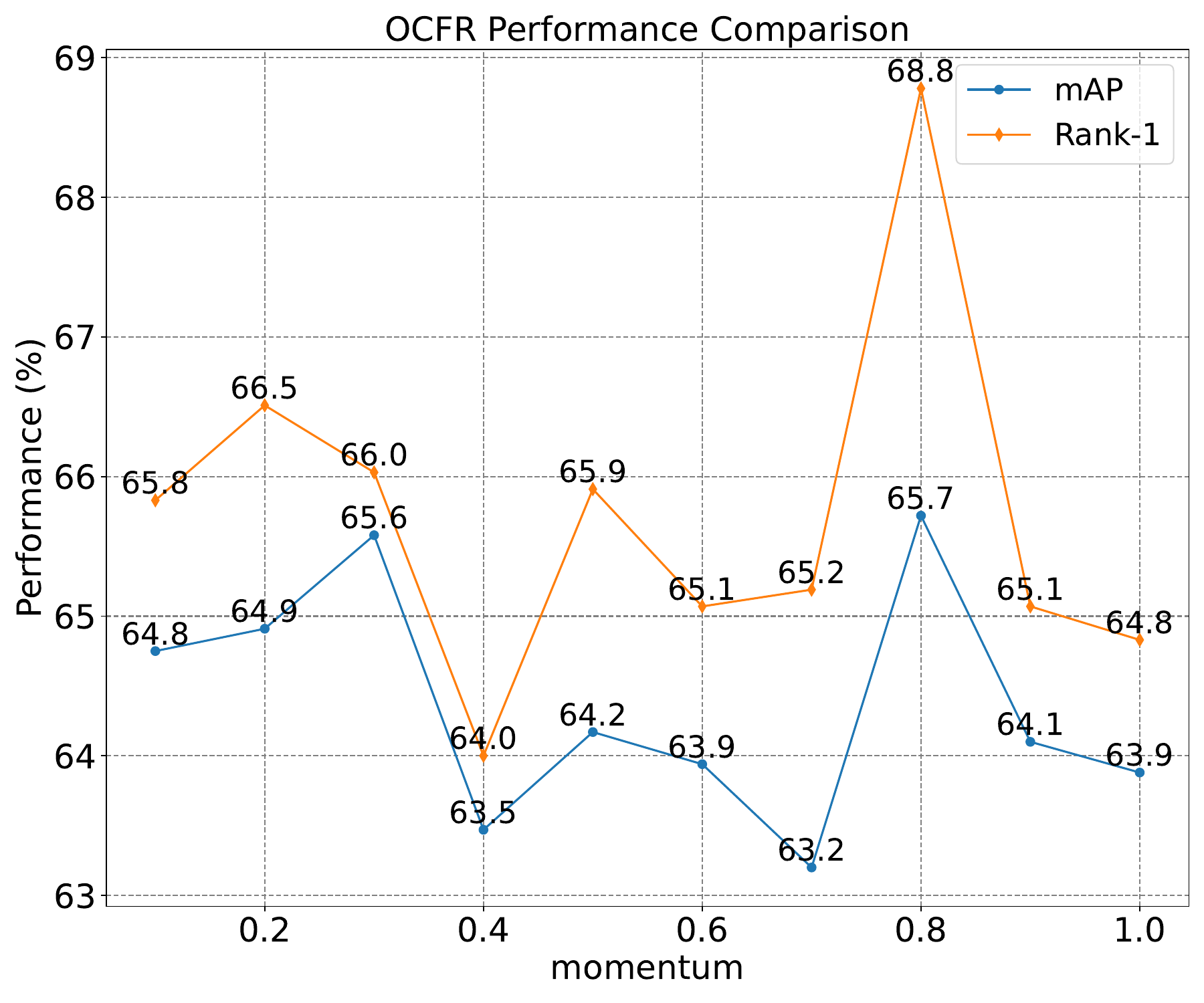}
  \caption{Effect of exponential parameter in OCFR.
  }
  \label{fig:ocfr}
\end{figure}
\\
\textbf{Effect of Exponential Parameter in OCFR.}
As shown in Fig. \ref{fig:ocfr}, with the OCFR parameter gradually increases from 0.1 to 0.8, the model shows a fluctuation in performance.
The mAP increases from 64.8\% at OCFR 0.1 to 65.7\% at OCFR 0.8.
Similarly, Rank-1 improves from 65.8\% to 68.8\%.
However, a slight performance decline is observed when OCFR is set to 1.0.
This indicates that a moderate momentum parameter in OCFR can enhance the model's ability to aggregate features with the same ID within each modality.
However, excessively large values may introduce noise, impacting overall performance.
\begin{figure}[t]
  \centering
  \includegraphics[width=1.000\linewidth]{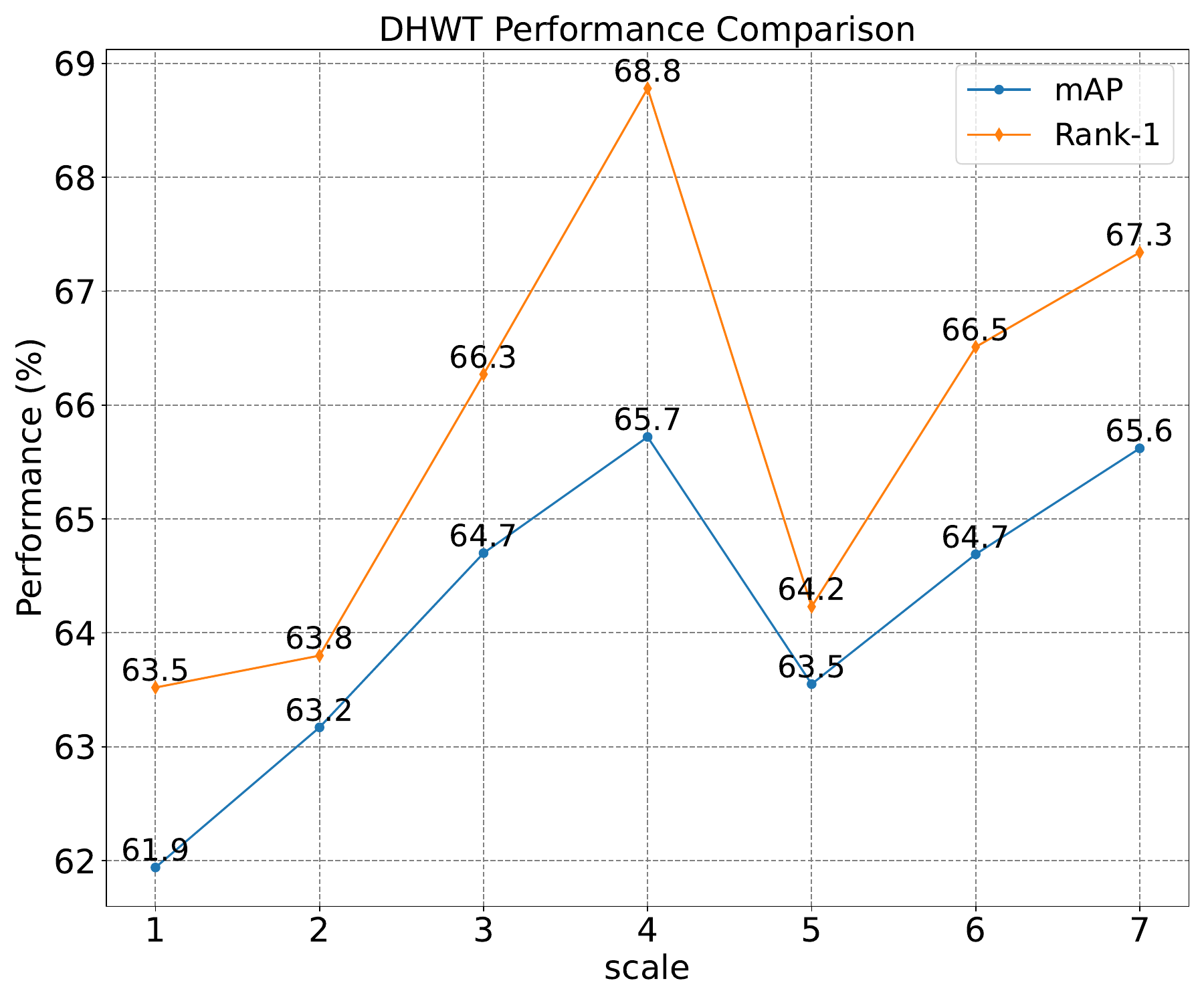}
  \caption{Effect of decomposition scales in DHWT.}
  \label{fig:dhwt}
\end{figure}
\\
\textbf{Effect of Decomposition Scales in DHWT.}
Fig. \ref{fig:dhwt} indicates that with the increase in decomposition levels, the overall metrics show an initial rise followed by a decline, reaching optimal results when the decomposition level is 4.
The results reveals the impact of DHWT decomposition scales on performance, showcasing an improvement in detail discernment with higher hierarchical scales.
As the hierarchy increases, the frequency-based selection demonstrates enhanced control over fine details in the images.
This suggests that a more intricate decomposition hierarchy contributes to better performance in capturing image details.
However, at excessively high levels, there is a risk of introducing irrelevant noise, resulting in a certain decline.
\begin{figure}[t]
  \centering
  \includegraphics[width=1.000\linewidth]{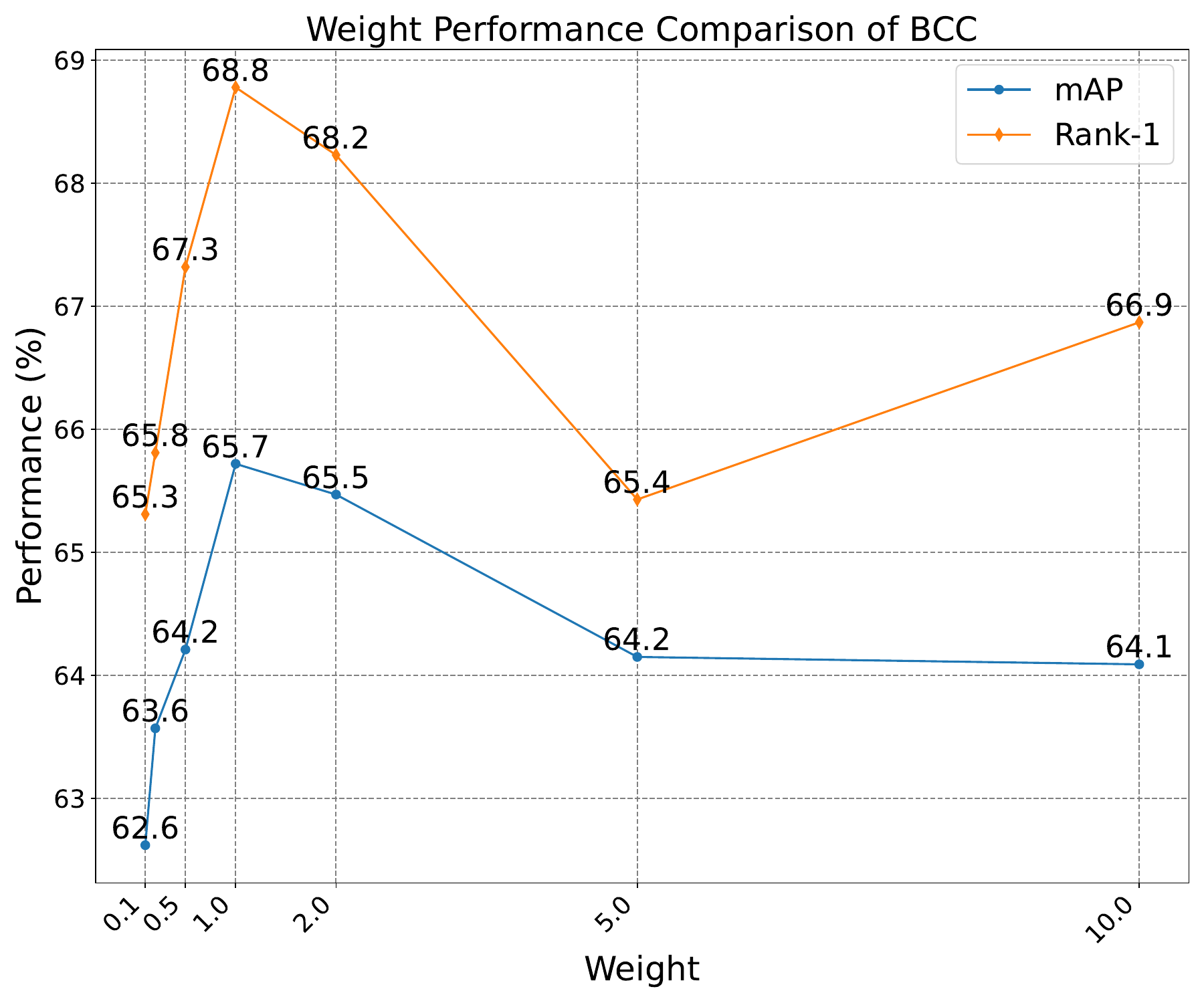}
  \caption{Effect of the weight of BCC loss.
  }
  \label{fig:bcc_weight}
\end{figure}
\\
\textbf{Effect of the Weight of BCC Loss.}
Fig. \ref{fig:bcc_weight} shows that mAP and Rank-1 achieve their peaks at a BCC loss weight of 1.
As the weight increases, performance gradually decreases, suggesting that an overly emphasized background consistency may lead to a decline in performance.
Therefore, finding a balance between dynamic alignment and background consistency is crucial.
\begin{figure}[t]
  \centering
  \includegraphics[width=1.000\linewidth]{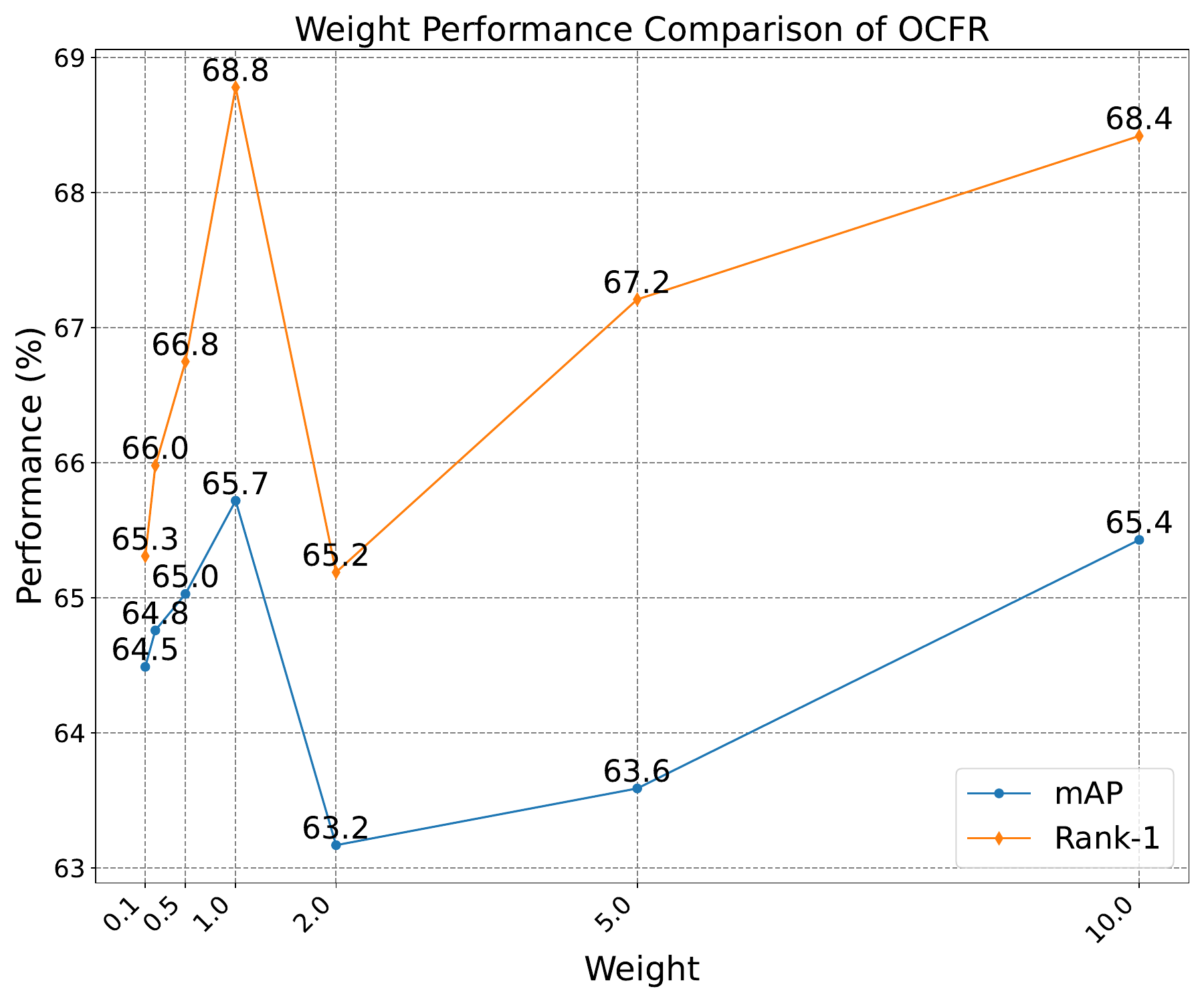}
  \caption{Effect of the weight of OCFR loss.
  }
  \label{fig:ocfr_weight}
\end{figure}
\begin{figure}[t]
  \centering
  \includegraphics[width=1.0\linewidth]{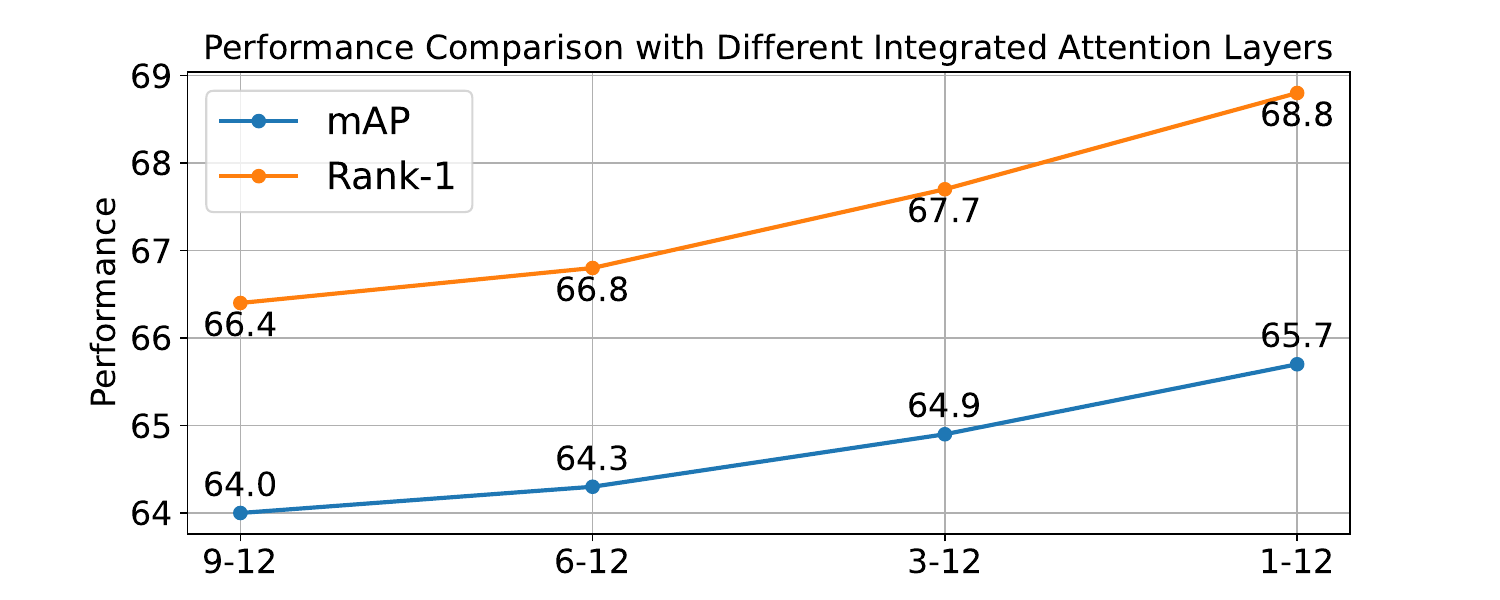}
   \caption{Comparison of different integrated attention layers.}
   \label{fig:attn}
\end{figure}
\\
\textbf{Effect of the Weight of OCFR Loss.}
Fig. \ref{fig:ocfr_weight} indicates that the model's optimal performance stems from a balanced emphasis on intra-modal feature alignment and inter-modal feature discrimination via the OCFR loss.
However, as the weight increases, there is an apparent excessive focus on inter-modal discrimination, resulting in reduced intra-modal alignment and overall performance.
Furthermore, certain weight values introduce model instability, leading to fluctuations in performance.
\\
\textbf{Effect of Integrated Attention Layers.}
In Fig. \ref{fig:attn}, we compare the performance of different integrated attention layers.
The results indicate that with all layers, our model achieves the best performance, demonstrating the importance of integrating multi-level attention for multi-modal representation learning.
\\
\textbf{Effect of Spatial-based Token Selection.}
Fig. \ref{fig:head} shows the impact of token numbers retained per head in spatial-based token selection.
As the number increases, the mAP and Rank-1 initially rise and then decline.
Meanwhile, the reserved tokens show a rapid increase, reaching 87.5\% when each head retains 8 tokens.
However, this leads to background interferences, resulting in poorer performance.
Notably, with two tokens per head, our model significantly improves the performance by capturing more object-centric parts in each modality.
\begin{figure}[t]
    \centering
    \resizebox{0.5\textwidth}{!}
	{
    \includegraphics[width=1.\linewidth]{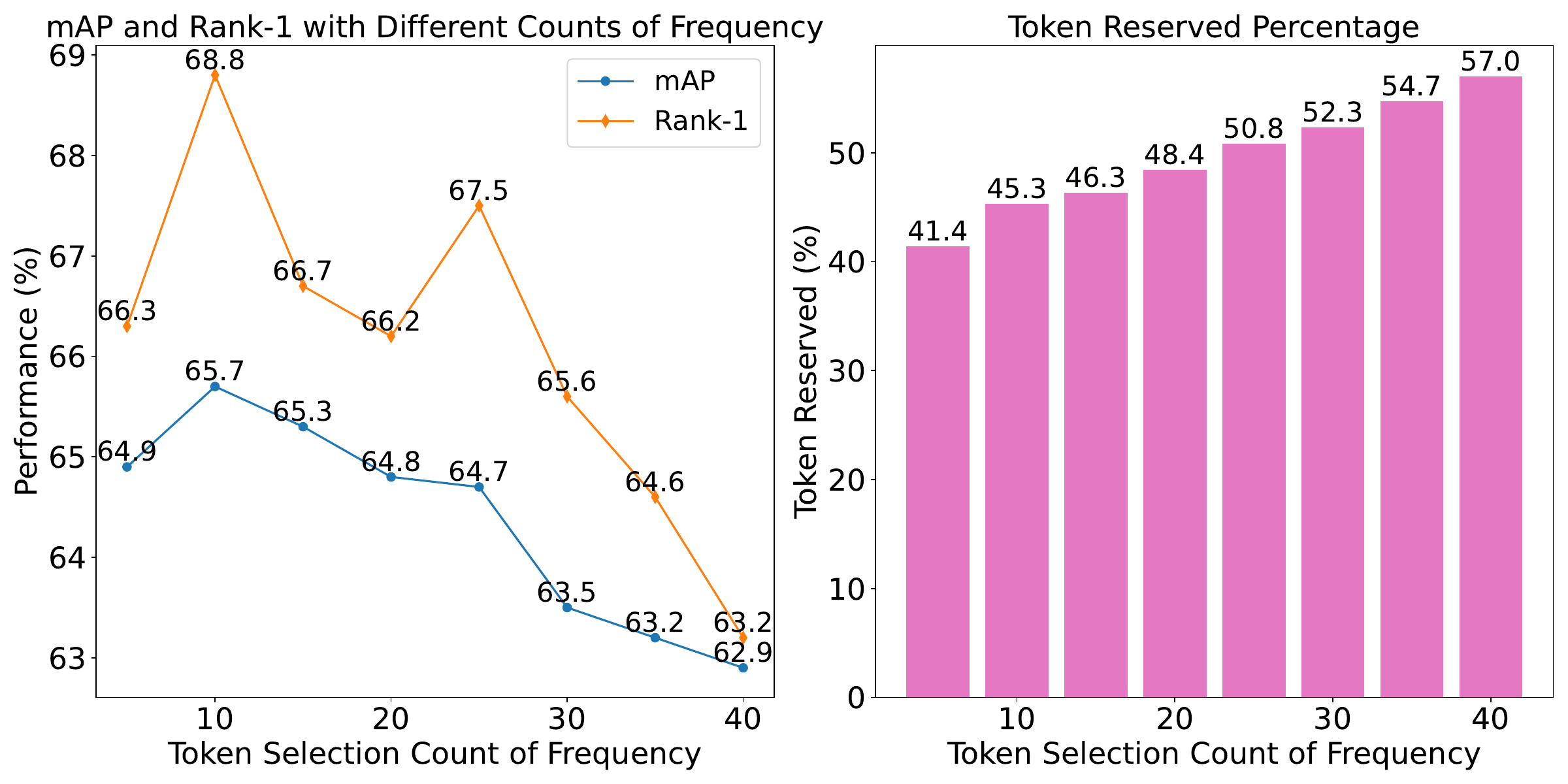}
    }
    \vspace{-2mm}
    \caption{Performance comparison of frequency-based selection.}
    \vspace{-4mm}
    \label{fig:fre}
\end{figure}
\\
\textbf{Effect of Frequency-based Token Selection.}
In Fig.~\ref{fig:fre}, we study how frequency-based token selection impacts the model.
Similar to Fig.~\ref{fig:head}, there is an initial increase followed by a decrease.
As the frequency-based tokens increase, the model's performance decline is more noticeable than with spatial-based selection.
This is because frequency-based tokens remain fixed, not adapting with learning.
Some salient tokens may introduce extra noise.
\subsection{More Ablations on Multi-modal Vehicle ReID}
\textbf{Parameter Analysis in EDITOR.}
In Tab. \ref{tab:params}, we present a parameter comparison of our method and other methods.
TransReID holds a large parameter count, while GraFT, leveraging a shared backbone for feature extraction, significantly reduces parameters.
In contrast, our EDITOR has comparable parameters to GraFT, but delivers much better results.
It even shows better results than TOP-ReID, which has a larger parameter count.
\begin{table}[h]
  \renewcommand\arraystretch{1.2}
  \setlength\tabcolsep{3pt}
  \centering
  \caption{Parameter comparison in our framework.}
\begin{tabular}{ccccc}
    \hline
    \multicolumn{1}{c}{\multirow{2}{*}{Methods}} &\multicolumn{1}{c}{\multirow{2}{*}{Params(M)}} &\multicolumn{2}{c}{RGBNT100}\\
    \cline{3-4}
    & & mAP & Rank-1\\
    \hline
  PCB~\cite{sun2018beyond}  &72.33 & 57.2 &83.5\\
  OSNet~\cite{zhou2019omni} &7.02& 75.0 &95.6 \\
  HAMNet~\cite{li2020multi} &  78.00 &74.5 &93.3 \\
  CCNet~\cite{zheng2022multi} &  74.60 & 77.2 &96.3\\
  GAFNet~\cite{guo2022generative} &  130.00 &74.4 &93.4\\
  \hline
  TransReID{*}~\cite{he2021transreid} & 278.23&75.6 &92.9\\
  UniCat{*}~\cite{crawford2023unicat}  & 259.02 &79.4&96.2\\
  GraFT{*}~\cite{yin2023graft} & 101.00 &76.6&94.3\\
  TOP-ReID{*}~\cite{wang2023top} & 324.53 &\underline{81.2}&\underline{96.4}\\
  \hline
  EDITOR{*}        & 118.55 & \textbf{82.1} &\textbf{96.4} \\
  \hline
  \end{tabular}
  \label{tab:params}
  \end{table}
  \\
  \textbf{Effect of Key Components on RGBNT100.}
In Tab. \ref{tab:ablation_vehicle}, we present a comprehensive performance comparison of various components on vehicle dataset RGBNT100.
We improve the baseline model (Model A) with more components.
Model B, incorporating HMA, demonstrates a 2.7\% increase in mAP, showcasing its effectiveness in aggregating multi-modal features.
Model C, introducing SFTS, achieves further improvement, highlighting the efficacy of object-centric token selection.
The integration of BCC (Model D) dynamically aligns multi-modal distributions, resulting in a substantial 1.5\% mAP enhancement compared to Model C.
Additionally, Model E enhances feature distribution compactness, leading to robust improvements.
The combination of all components in EDITOR (Model F) achieves optimal performance, verifying the effectiveness of our methods on vehicle datasets.
\begin{table}[t]
  \centering
  \renewcommand\arraystretch{1.12}
  \setlength\tabcolsep{3pt}
  \caption{Performance comparison with different components.}
  \resizebox{0.42\textwidth}{!}
{
  \begin{tabular}{c|cc|cc|cccc}
      \hline
      &\multicolumn{2}{c|}{Module} &\multicolumn{2}{c|}{Loss}       &\multicolumn{4}{c}{RGBNT100} \\
      & SFTS& HMA& BCC& OCFR      & mAP & R-1 & R-5 & R-10 \\
      \hline
      A & \ding{53}& \ding{53}& \ding{53}& \ding{53}& 75.1 &93.4 &95.0 &95.8\\
      B & \ding{53}& \ding{51}& \ding{53}& \ding{53}& 77.8 & 94.0 &95.1 & 96.0 \\
      C & \ding{51}& \ding{51}& \ding{53}& \ding{53}& 79.1 & 94.3 & 95.3 & 96.1 \\
      D & \ding{51}& \ding{51}& \ding{51}& \ding{53}& 80.6 &	95.5 & 96.4	&97.2\\
      E & \ding{51}& \ding{51}& \ding{53}& \ding{51}& 80.4 & 94.8 & 95.5 & 96.3 \\
      F & \ding{51}& \ding{51}& \ding{51}& \ding{51}& \textbf{82.1}        & \textbf{96.4}           & \textbf{96.9}         & \textbf{97.4}             \\
      \hline
  \end{tabular}
  }
  \label{tab:ablation_vehicle}
\end{table}
\section{Visualization}
\begin{figure}[t]
  \centering
  \includegraphics[width=1.\linewidth]{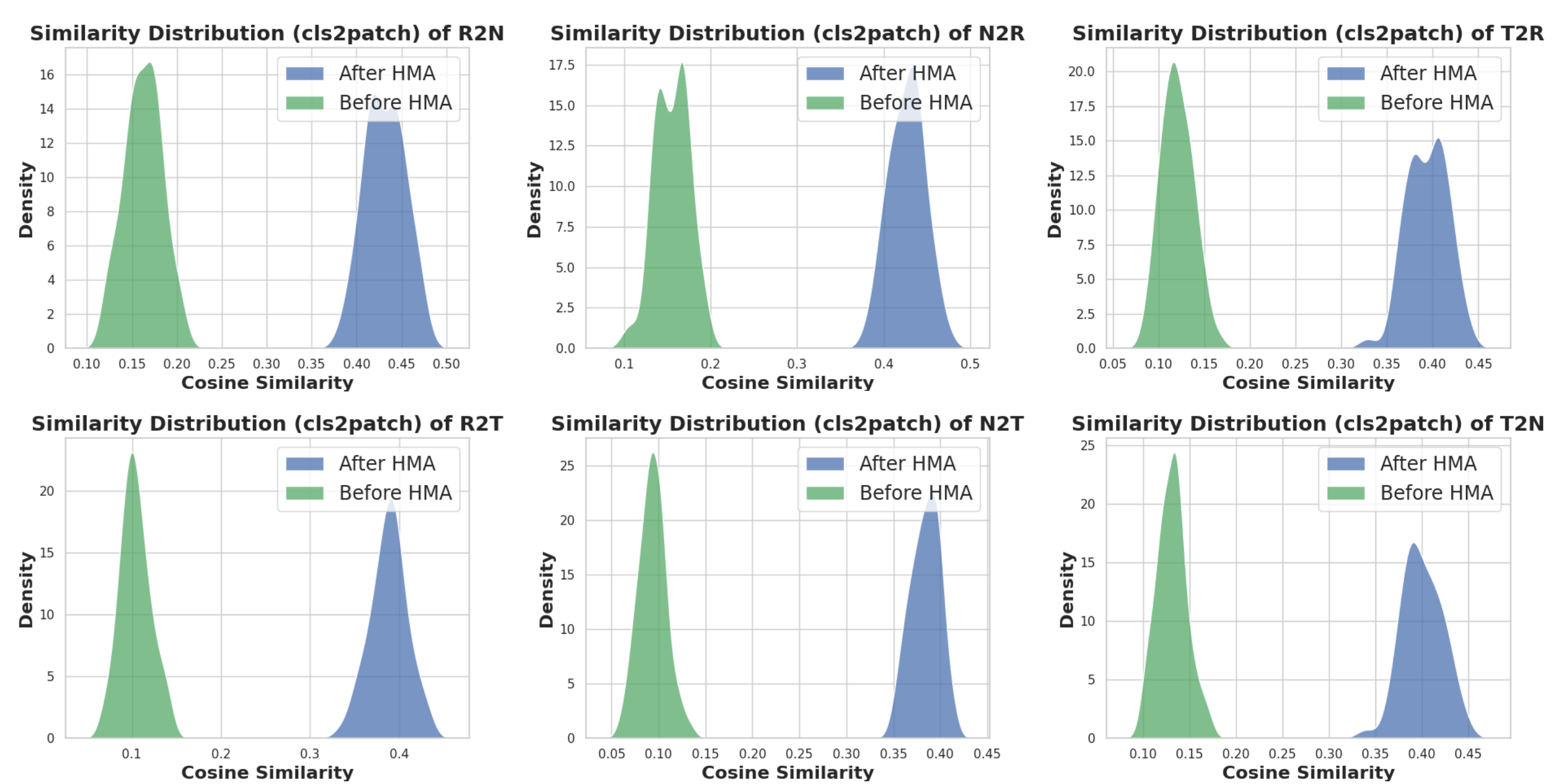}
  \caption{Alignment visualization in HMA with all modalities.}
  \label{fig:more_HMA}
\end{figure}
\textbf{More Visualizations of HMA on Feature Alignment.}
In Fig. \ref{fig:more_HMA}, we employ the cosine similarity distribution between class tokens of different modalities, examining the impact before and after HMA.
The results highlight the notable improvement in aligning class tokens with patch tokens across modalities after HMA, showcasing the efficacy of HMA in enhancing the feature alignment and aggregation for multi-modal representations.
\begin{figure}[t]
  \centering
  \includegraphics[width=1.\linewidth]{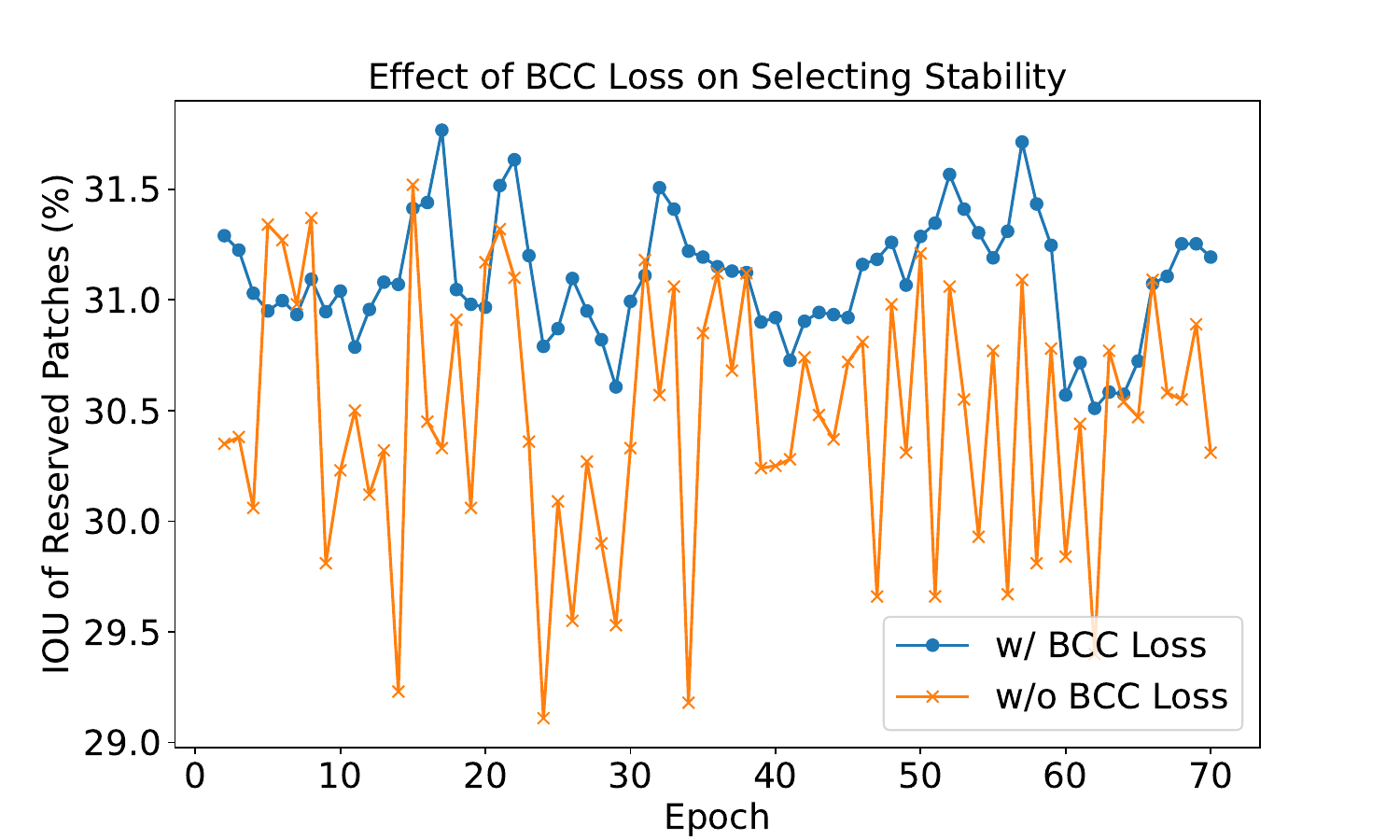}
  \caption{Selecting stability of reserved patches.}
  \label{fig:stable}
  \vspace{-4mm}
\end{figure}
\begin{figure*}[t]
  \centering
  \includegraphics[width=1.000\linewidth]{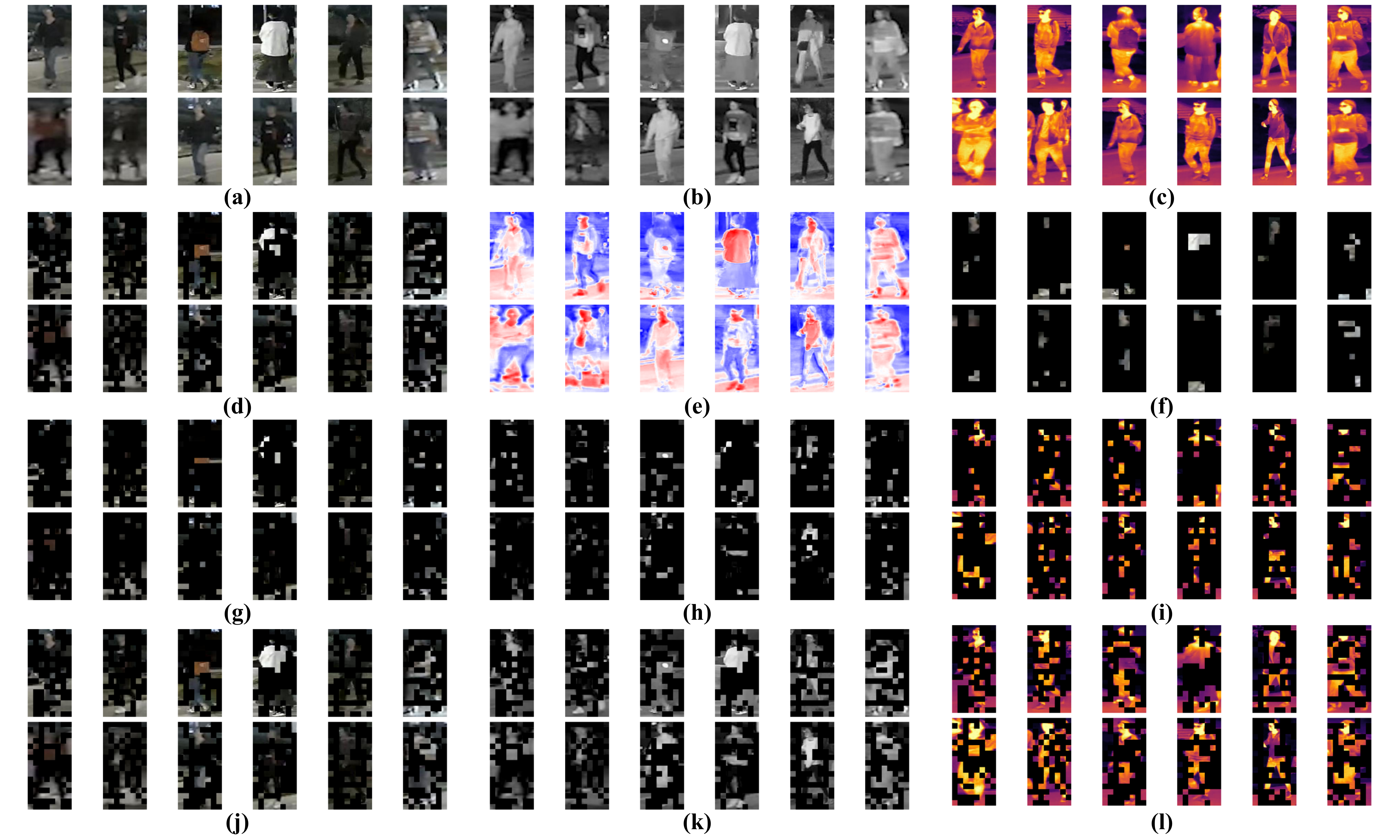}
  \caption{Visualization of selected tokens at different stages (Person).
  (a) RGB images;
  (b) NIR images;
  (c) TIR images;
  (d) Spatial-based token selection;
  (e) DHWT effect;
  (f) Frequency-based token selection;
  (g-i) Spatial-based token selection from RGB/NIR/TIR;
  (j-l) Final tokens for RGB/NIR/TIR.
  Note that we project the selected tokens back to the corresponding image regions.
  }
  \label{fig:select_p}
\end{figure*}
\begin{figure*}[t]
  \centering
  \includegraphics[width=1.000\linewidth]{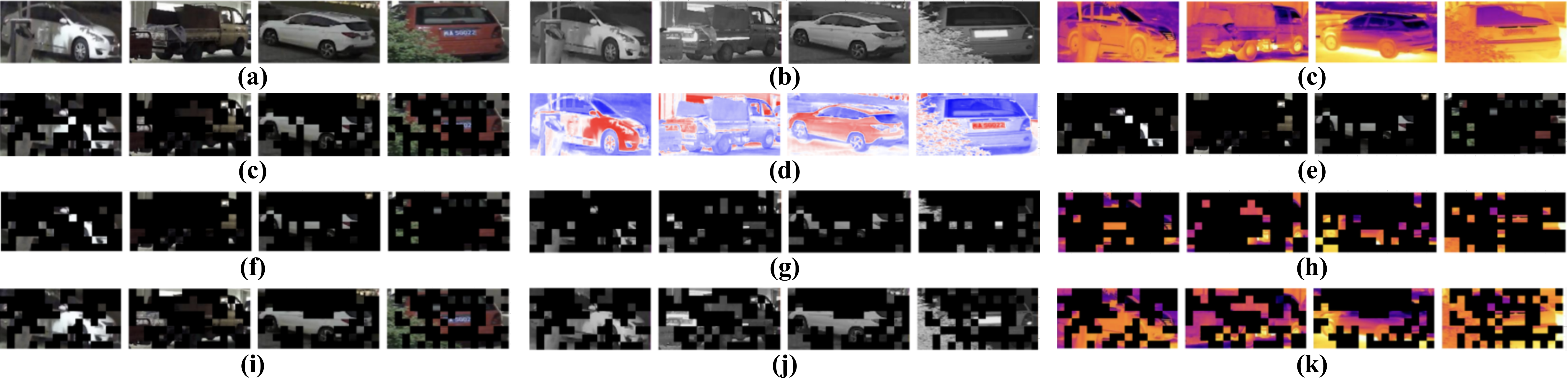}
  \caption{Visualization of selected tokens at different stages (Vehicle).
  (a) RGB images;
  (b) NIR images;
  (c) TIR images;
  (d) Spatial-based token selection;
  (e) DHWT effect;
  (f) Frequency-based token selection;
  (g-i) Spatial-based token selection from RGB/NIR/TIR;
  (j-l) Final tokens for RGB/NIR/TIR.
  Note that we project the selected tokens back to the corresponding image regions.
  }
  \label{fig:select_v}
\end{figure*}
\\
\textbf{Stability of the BCC Loss.}
The BCC loss stabilizes the selection process by aligning background features from different modalities.
In Fig. \ref{fig:stable}, we depict the Intersection over Union (IOU) of the retained patches between adjacent epochs, comparing the scenarios with and without the BCC loss throughout the training process.
It is evident that the BCC loss introduces a noticeable smoothing effect on the token selection process across the entire dataset, maintaining a more stability of certain patches.
\\
\textbf{Selected Tokens at Different Stages.}
In Fig. \ref{fig:select_p} and Fig. \ref{fig:select_v}, we visualize the selected tokens at different stages on the person ReID dataset RGBNT201 and the vehicle dataset RGBNT100, respectively.
Taking RGBNT201 as an example, in Fig. \ref{fig:select_p}, the top row displays the input images from various modalities, revealing distinct details.
In Fig. \ref{fig:select_p}(e), we present the results after performing DHWT for collaborative transformation, highlighting significant areas corresponding to object-centric regions.
Fig. \ref{fig:select_p}(g)-(i) illustrate the image regions corresponding to the selected tokens by individual modalities, emphasizing the substantial differences in the focused areas.
Through modality union, we capture a diverse range of detailed regions, yielding the composite result shown in Fig. \ref{fig:select_p}(d).
In Fig. \ref{fig:select_p}(d), we can observe the effect of using spatial attention for selection, already effectively capturing most crucial areas.
By incorporating the selection of other object-centric regions in Fig. \ref{fig:select_p}(f), we obtain the final selection result as shown in \ref{fig:select_p}(j)-(l).
In Fig. \ref{fig:select_p}(l), we clearly observe that essential features of the human body are well-preserved.
Similar results can be observed in Fig. \ref{fig:select_v} on the vehicle dataset.
These visualizations fully validate the effectiveness of our EDITOR.

\end{document}